\documentclass[11pt]{article}

\usepackage[]{acl}

 \usepackage{microtype}
\usepackage{tabularx}
\usepackage{booktabs}
\usepackage{multirow}
\usepackage[table]{xcolor}
\usepackage{colortbl}
\usepackage{caption}
\usepackage{cuted}

\definecolor{sectiongray}{RGB}{230,230,230}
\definecolor{lightyellow}{RGB}{255,240,245}
\definecolor{lightblue}{RGB}{225,238,248}
\usepackage[english,bidi=default]{babel} 
\babelfont{rm}{TeXGyreTermesX} 
\babelprovide[import]{hindi}
\babelfont[*devanagari]{rm}{Lohit Devanagari}
\babelprovide[import]{arabic}
\babelfont[*arabic]{rm}{Noto Sans Arabic}

\usepackage{needspace}

\usepackage{amsmath, amssymb, amsfonts}
\usepackage{subcaption}
\usepackage{booktabs}
\usepackage{multirow}
\usepackage{siunitx}
\usepackage{enumitem}
\usepackage{algorithm}
\usepackage{algpseudocode}
\usepackage{tikz}
\usetikzlibrary{arrows.meta, positioning, shapes.geometric, shapes.misc, calc, fit}
\usepackage{listings}
\usepackage{xcolor}
\usepackage[most]{tcolorbox}
\usepackage{pifont}
\usepackage{xspace}
\usepackage[strings]{underscore}
\usepackage{fontspec}
\usepackage{cuted}
\usepackage{caption}
\usepackage{tabularx}
\newfontface\banglafont[Script=Bengali]{Kalpurush.ttf}
\newcommand{\bn}[1]{{\banglafont #1}}

\definecolor{PromptHeader}{HTML}{2B2B2B}
\definecolor{PromptBorder}{HTML}{4A6FA5}
\definecolor{PromptBody}{HTML}{FFFFFF}
\definecolor{PromptText}{HTML}{000000}
\definecolor{PromptBlue}{HTML}{EAF2FB}
\definecolor{ResultHeader}{HTML}{2B2B2B}
\definecolor{ResultBorder}{HTML}{4A8F6D}
\definecolor{ResultBody}{HTML}{FFFFFF}
\definecolor{ResultGreen}{HTML}{EAF6EE}

\newtcolorbox{promptbox}[1][]{
  enhanced,
  colback=PromptBlue,
  colframe=PromptBorder,
  coltitle=white,
  fonttitle=\bfseries,
  title=#1,
  boxrule=0.8pt,
  arc=2mm,
  left=2mm,
  right=2mm,
  top=1mm,
  bottom=1mm
}

\newtcolorbox{resultbox}[1][]{
  enhanced,
  colback=ResultGreen,
  colframe=ResultBorder,
  coltitle=white,
  fonttitle=\bfseries,
  title=#1,
  boxrule=0.8pt,
  arc=2mm,
  left=2mm,
  right=2mm,
  top=1mm,
  bottom=1mm
}

\newcommand{\dataset}{\textsc{UnrestSent200K}}
\newcommand{\labels}{\{\text{positive},\,\text{neutral},\,\text{negative}\}}

\newcommand{\blfootnote}[1]{%
  \begingroup
  \renewcommand\thefootnote{}\footnote{#1}%
  \addtocounter{footnote}{-1}%
  \endgroup
}
\title{Can LLMs Follow the Pulse of a Crisis? Evaluating Crisis Sentiment in Bangladesh's July Uprising}

\author{
\textbf{Md.\ Samiul Alim}$^{1,*}$ \quad
\textbf{Mahir Shahriar Tamim}$^{1,*}$ \quad
\textbf{Tanvir Ahmed Khan}$^{1}$ \quad
\textbf{Sharjil Khan}$^{1}$ \\
\textbf{Rafia Ferdous Duti}$^{1}$ \quad
\textbf{Shahriyar Zaman Ridoy}$^{1}$ \quad
\textbf{Mohammad Ali Moni}$^{2,\dagger}$ \\[3pt]
$^{1}$North South University, Dhaka, Bangladesh \\
$^{2}$Charles Sturt University, Australia \\[3pt]
{\scriptsize samiul.alim01@northsouth.edu, \enspace
mahir.tamim@northsouth.edu, \enspace
khan.tanvir01@northsouth.edu,} \\
{\scriptsize sharjil.khan@northsouth.edu, \enspace
rafia.duti@northsouth.edu, \enspace
shahriyar.zaman01@gmail.com, \enspace
mmoni@csu.edu.au}
}

\begin{document}
\raggedbottom
\maketitle
\blfootnote{$^{*}$\,These authors contributed equally to this work.}
\blfootnote{$^{\dagger}$\,Corresponding author.}

\begin{strip}
    \centering
    \includegraphics[width=0.24\textwidth]{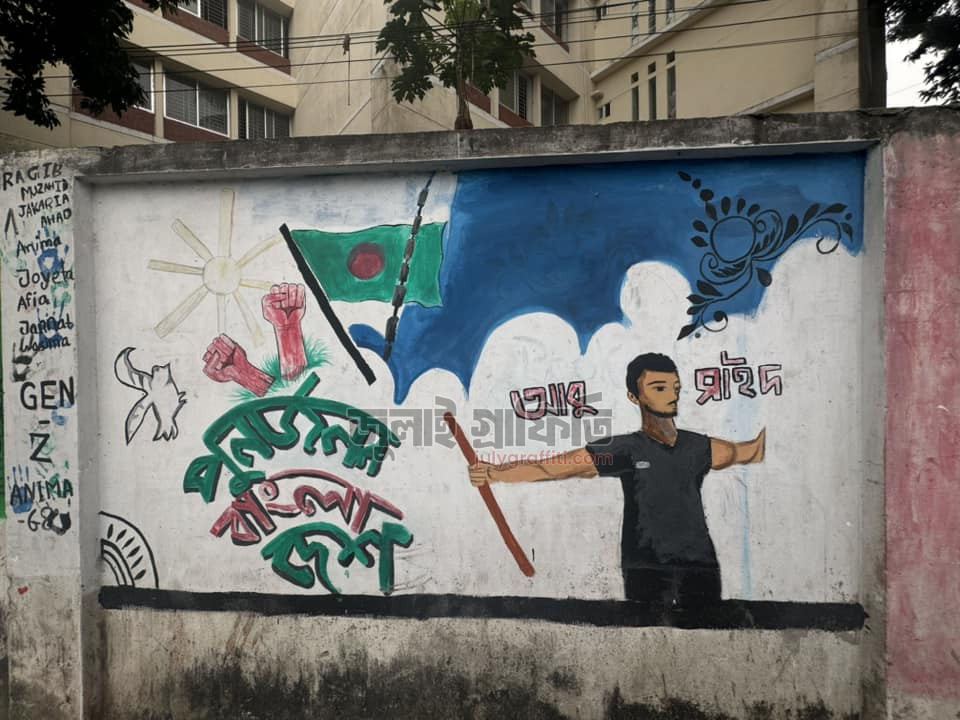}\hfill
    \includegraphics[width=0.24\textwidth]{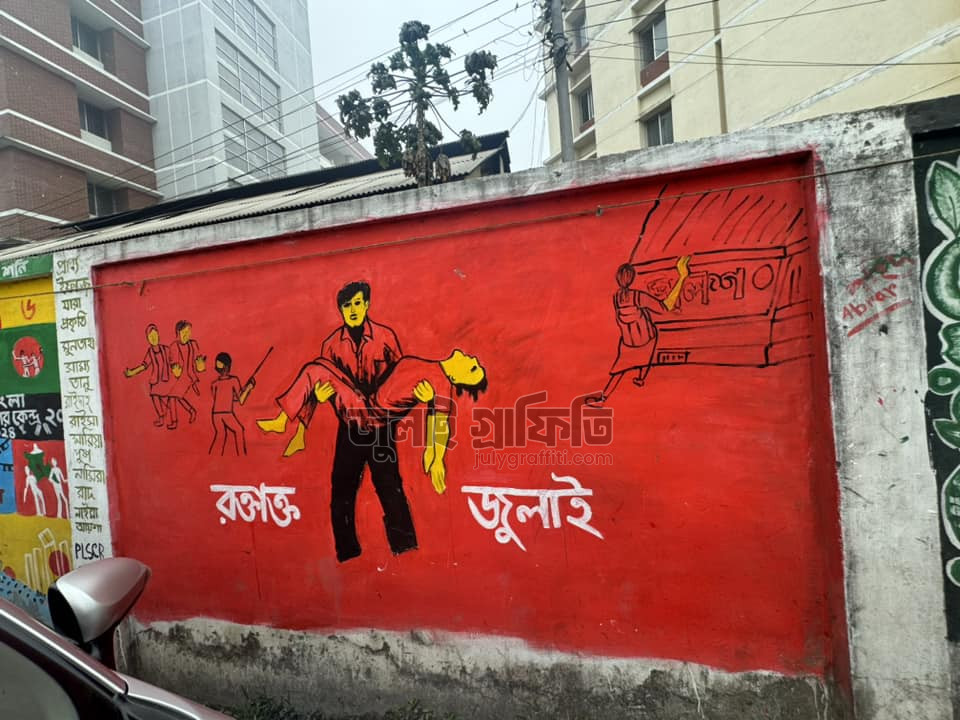}\hfill
    \includegraphics[width=0.24\textwidth]{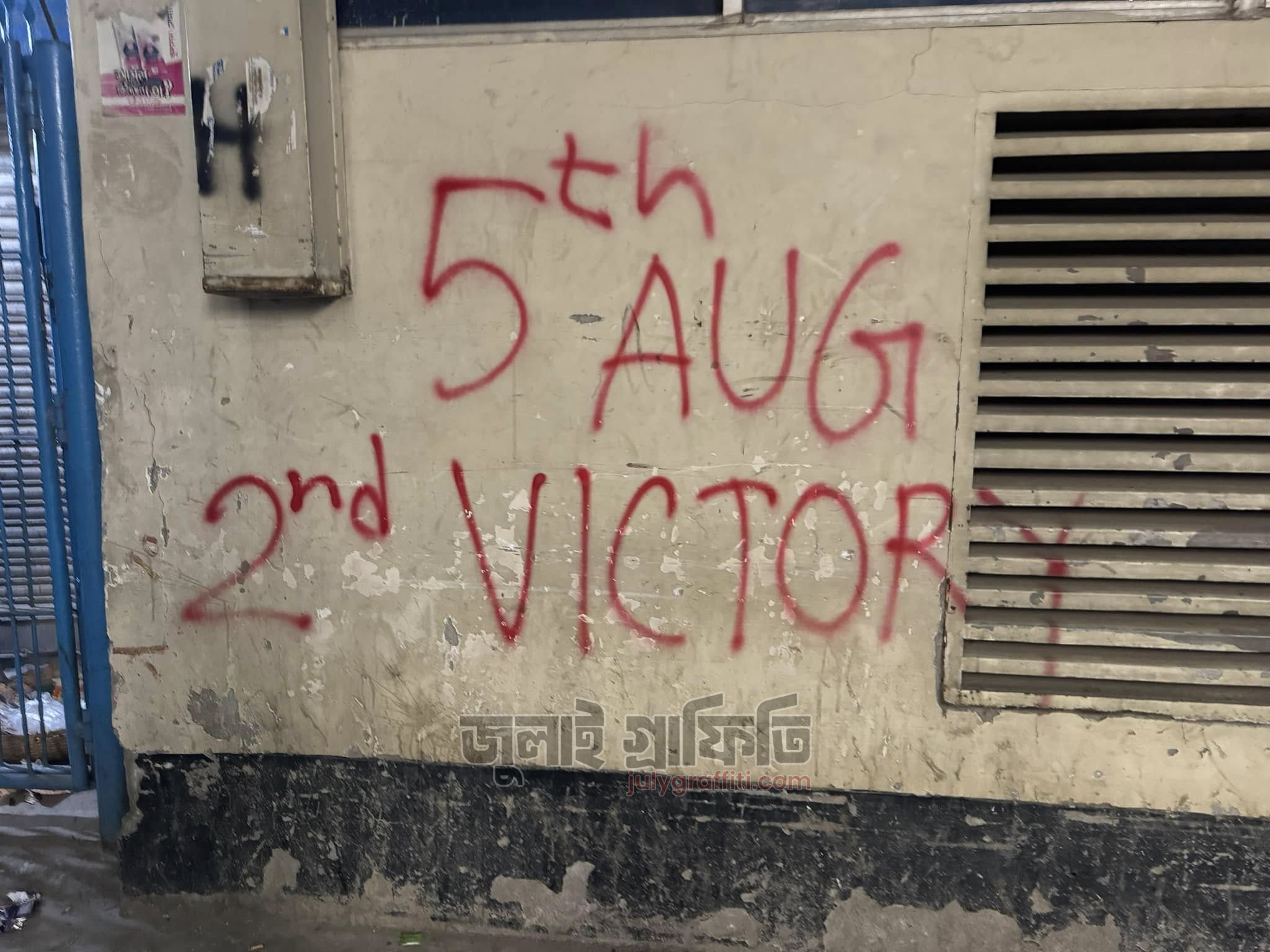}\hfill
    \includegraphics[width=0.24\textwidth]{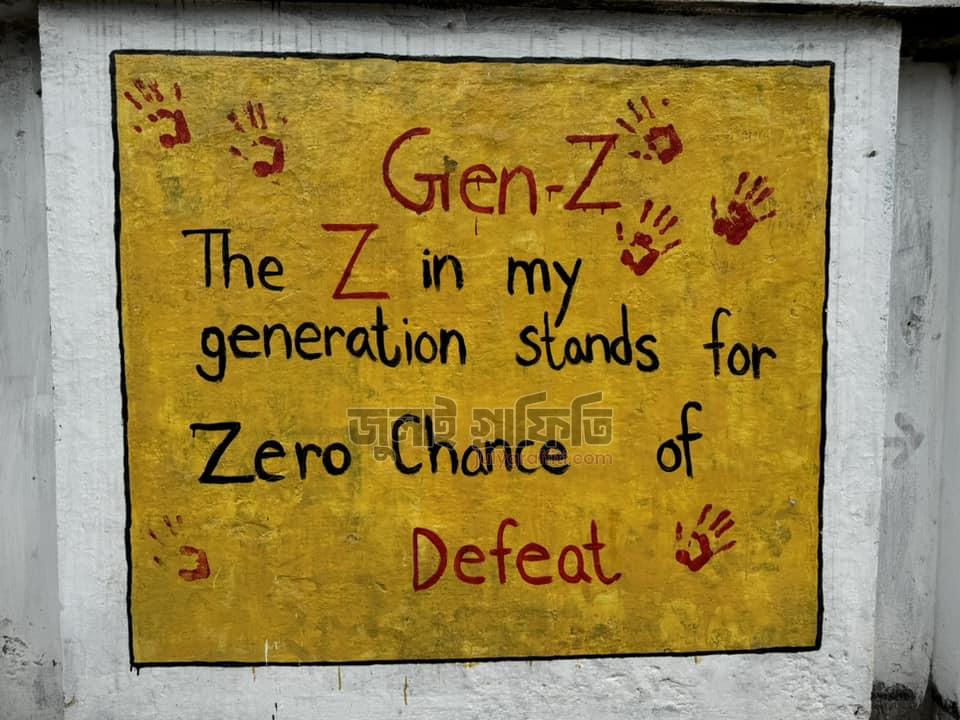}
\captionof{figure}{Graffiti from the July 2024 student-led uprising in Bangladesh, illustrating public expressions of resistance, solidarity, generational identity, and political sentiment captured on urban walls.}
    \label{fig:graffiti}
\end{strip}

\begin{abstract}
Crisis sentiment analysis is especially challenging for low-resource languages such as Bangla, where language, context, and public reaction shift rapidly. We introduce \textsc{UnrestSent200K}, a Bangla crisis sentiment dataset with $\approx$200K Facebook and YouTube comments from the July--August 2024 Bangladesh uprising. The dataset covers five event-aligned phases, from early escalation and internet blackout to regime transition and a later flood crisis. Each comment is linked to its parent post, enabling evaluation with and without discourse context. All comments are annotated through a fully human process involving 14 native Bangla-speaking annotators and senior validation, achieving substantial agreement ($\kappa = 0.73$, $\alpha = 0.71$) and 94.2\% blind-audit agreement. We benchmark fine-tuned encoders, prompted LLMs, and LoRA-tuned LLMs. Results show that parent-post context consistently improves performance, while temporal shift across phases causes large performance drops. Strong LLMs perform well, but still struggle with sarcasm, implicit political references, and phase-dependent meaning. \dataset{} provides a benchmark for studying context-aware and temporally robust sentiment analysis in low-resource crisis discourse. \textsc{UnrestSent200K} is available at \url{https://sami0055.github.io/UNRESTSENT200K/}
\end{abstract}

\section{Introduction}

During a political crisis, public language can change almost overnight. A short social-media comment that appears positive at one moment may sound sarcastic, fearful, or critical after a major event. Its meaning may also depend on the news post to which it responds. Yet most sentiment benchmarks treat language as stable over time, evaluate each utterance in isolation, and focus mainly on English~\cite{Socher2013,Maas2011,Go2009}. These assumptions are especially fragile in crisis settings, where sentiment systems are increasingly used to support situational awareness, journalism, and policy decisions~\cite{Tufekci2017,steinert2017spontaneous}. The central problem is therefore not only a lack of data. It is also a lack of evaluation settings that show whether models can follow a rapidly changing public conversation.

The July--August 2024 Bangladesh uprising~\cite{RANA2026100341} offers a rare opportunity to study this problem (Figure~\ref{fig:graffiti}). In only seven weeks, public discussion on Facebook and YouTube moved through five distinct phases: early mobilisation, a nationwide internet blackout, regime collapse, political transition, and an overlapping flood crisis. The language, platforms, and broad user population remained largely the same, but the social and political context changed sharply. This creates a natural stress test: can a model trained during one phase still understand sentiment in the next, and can it interpret a comment without seeing the post that prompted it?

For Bangla, these questions have been difficult to answer. Existing resources such as SentNoB, SentiGOLD, and BnSentMix~\cite{SentNoB2021,SentiGOLD2023,Alam2024BnSentMix} have made important progress, but they are mostly static, decontextualised, and focused on consumer or general social-media text. They do not divide a single unfolding event into meaningful time periods, link comments to their parent posts, or support controlled comparisons across phases. As a result, they cannot reveal how quickly model performance changes as an event develops. To fill this gap, we introduce \dataset{}, the largest Bangla sentiment dataset to date, with approximately 200K comments ($2.8\times$ SentiGOLD), as summarised in Table~\ref{tab:dataset_comparison}. Every comment is timestamped, assigned to one of five event-aligned phases, and paired with its parent post. The labels were produced entirely by 14 native Bangla-speaking undergraduate annotators who had first-hand exposure to the events. Our five-stage \textbf{Pilot Label Annotation (PiLA)} framework includes calibration, double annotation, senior adjudication, and an independent blind audit. It achieves substantial agreement (Cohen's $\kappa=0.73$ and Krippendorff's $\alpha=0.71$) and 94.2\% agreement in the blind audit.

Our experiments tell a consistent story. We evaluate four fine-tuned encoders (BanglaBERT, mBERT, XLM-R, and BanglaElectra), 15 zero- and few-shot LLMs, and two LoRA-tuned LLMs. Giving an encoder the parent post improves performance by 7--11 percentage points, showing that context is often essential rather than optional. However, context alone does not solve the problem: training on earlier phases and testing on later ones causes drops of 19--28 F1 points. Sentiment changes occur in sharp, event-linked phases rather than along a smooth trend. GPT-4o-mini, the strongest prompted model, reaches approximately 76\% few-shot accuracy but shares several failure modes with the encoders. LoRA-tuned Gemma-4B performs best overall at 78.4\% accuracy, yet even this result remains below 80\%. Together, these findings show that larger models and stronger prompting help, but do not remove the need for contextual grounding and temporal adaptation.
\paragraph{Contributions.}
\begin{itemize}[leftmargin=*,itemsep=2pt]

\item \textbf{A large Bangla crisis benchmark.}
We introduce \dataset{}, containing approximately 200K comments from the July--August 2024 Bangladesh uprising.

\item \textbf{A context- and time-aware design.}
Each comment is linked to its parent post and an event phase, enabling direct tests of contextual understanding and temporal robustness.

\item \textbf{Careful human annotation.}
PiLA provides a five-stage process with native Bangla-speaking annotators, senior adjudication, and an independent blind audit.

\item \textbf{Evidence across model families.}
Experiments with fine-tuned encoders, prompted LLMs, and LoRA-tuned LLMs show clear gains from parent-post context and large losses under temporal shift.

\end{itemize}

\section{Related Work}
\label{sec:related}

\textbf{Sentiment benchmarks and the low-resource gap.}
Sentiment analysis is driven by large static English benchmarks~\cite{Socher2013,Maas2011,McAuley2015,Go2009,HuLiu2004}. In Bangla, SentNoB~\cite{SentNoB2021} introduced noisy multi-domain social-media comments (15K), SentiGOLD~\cite{SentiGOLD2023} scaled to 70K with fine-grained labels, and BnSentMix~\cite{Alam2024BnSentMix} targeted code-mixed text. Pretrained Bangla encoders~\cite{Bhattacharjee2022BanglaBERT} and recent task-specific resources~\cite{ANUBHUTI2025,BanglaMUSE2026,BanglaSentNet2026,BANGLA_ABSA2024,BanglaECommerce2025,BanglaEmotion2024} have raised the ceiling, but all are static, decontextualised, and consumer-oriented; none is anchored to a single event whose internal phases can be used to probe within-episode shift~\cite{JulyRevolutionSentiment2025,SocioPoliticalEventsSurvey2024}.

\noindent\textbf{Temporal and contextual robustness.}
Prior work has studied concept drift and temporal generalisation in social-media classification~\cite{Tufekci2017,steinert2017spontaneous}, but often across years, topics, or domains, where temporal change is confounded with demographic and topical shifts. In contrast, \dataset{} isolates temporal shift within a single seven-week crisis, while keeping language, platform, and population largely fixed. It also operationalises contextual grounding by pairing each comment with its parent post, enabling direct comment-only versus post+comment evaluation.

\noindent\textbf{Human annotation for sensitive discourse.}
Politically sensitive crisis discourse requires annotators with strong linguistic and cultural grounding. Following best practices in corpus annotation~\cite{pustejovsky-2012-role,sabou-etal-2014-corpus,mukta2021comprehensive}, we use a fully human five-stage PiLA protocol with guideline piloting, double annotation, senior adjudication, and blind audit. The process involved 14 native-Bangla annotators with first-hand familiarity with the events and achieved substantial reliability: $\kappa = 0.73$, $\alpha = 0.71$, and 94.2\% blind-audit agreement. Thus, \dataset{} contributes not only a benchmark, but also a replicable annotation protocol for low-resource, politically sensitive NLP.

\begin{table}[t]
\centering
\scriptsize
\caption{Comparison of Bangla social-media datasets across sentiment, hate-speech, and offensive-language benchmarks.}
\label{tab:dataset_comparison}

\setlength{\tabcolsep}{2.5pt}

\begin{tabular}{lccc}
\toprule
\textbf{Dataset} & \textbf{Size} & \textbf{Task} & \textbf{Domain} \\
\midrule

SentNoB \cite{SentNoB2021}
& 15K 
& Sent. 
& Multi \\

SentiGOLD \cite{SentiGOLD2023}
& 70K 
& Sent. 
& Multi \\

BnSentMix \cite{Alam2024BnSentMix}
& 20K 
& Sent. 
& Code-mix \\

BD-SHS \cite{islam2022bdshs}
& 50K 
& HS 
& Social \\

Bengali Tweets \cite{das-etal-2022-hate-speech}
& 10K 
& Hate 
& Code-mix \\

TB-OLID \cite{raihan-etal-2023-offensive}
& 5K 
& Offensive 
& Transliterated \\

BanTH \cite{haider-etal-2025-banth}
& 37K 
& HS 
& Transliterated \\

BIDWESH \cite{fayaz2025bidweshbanglaregionalbased}
& 9K 
& HS 
& Dialectal \\

BanglaMultiHate \cite{hasan2025llmbasedmultitaskbanglahate}
& 50K 
& HS 
& YouTube \\

\midrule

\textbf{\dataset{} (Ours)} 
& \textbf{$\approx$200K} 
& \textbf{Sent.} 
& \textbf{Crisis} \\

\bottomrule
\end{tabular}

\vspace{-0.2cm}
\end{table}


\begin{figure*}[t]
    \centering    \includegraphics[width=\textwidth,height=0.50\textheight,keepaspectratio]{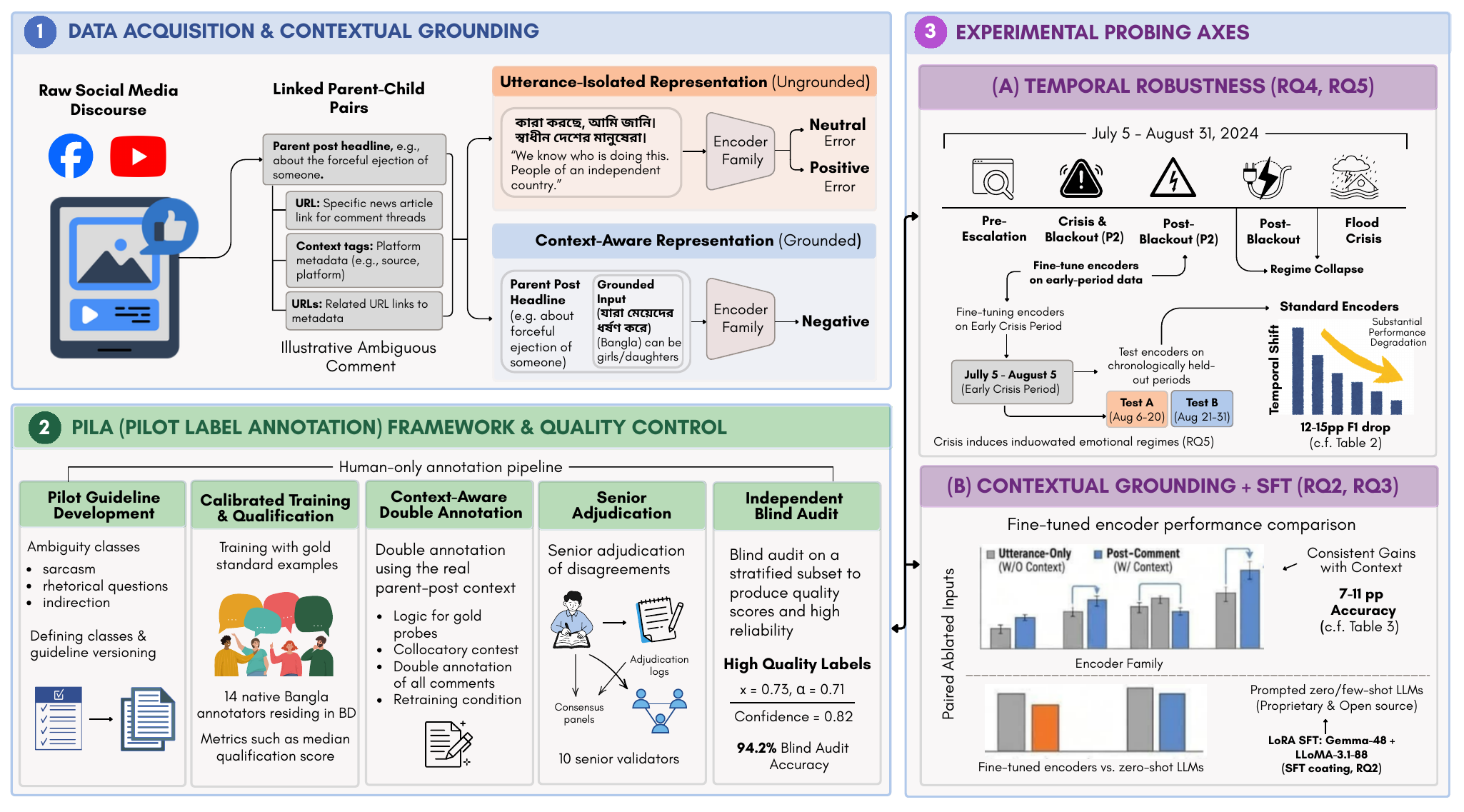}
\caption{
Overview of the \textsc{UnrestSent200K} construction and evaluation framework. \textbf{(1) Data acquisition and contextual grounding:} public Bangla crisis discourse is collected from Facebook and YouTube, comments are linked to their parent posts and metadata, and utterance-only and context-aware inputs are constructed. \textbf{(2) PiLA annotation and quality control:} a five-stage, fully human workflow covers guideline development, annotator calibration, context-aware double annotation, senior adjudication, and an independent blind audit. \textbf{(3) Experimental probing axes:} the benchmark evaluates temporal robustness across event phases and the effects of contextual grounding and supervised fine-tuning across encoder and LLM families.
}
    
    \label{fig:annotation_overview}
\end{figure*}

\begin{table*}[t]
\centering
\caption{Stratified 70/15/15 train/validation/test split of \dataset{} by event-aligned phase and sentiment class. Splits are jointly stratified on phase and sentiment to preserve the class and temporal distribution within each partition.}
\label{tab:split_phase_sentiment}
\small
\setlength{\tabcolsep}{4pt}
\renewcommand{\arraystretch}{1.05}
\resizebox{\textwidth}{!}{%
\begin{tabular}{ll rrr rrr rrr r}
\toprule
& & \multicolumn{3}{c}{\textbf{Train (70\%)}} & \multicolumn{3}{c}{\textbf{Validation (15\%)}} & \multicolumn{3}{c}{\textbf{Test (15\%)}} & \\
\cmidrule(lr){3-5} \cmidrule(lr){6-8} \cmidrule(lr){9-11}
\textbf{Phase} & \textbf{Period} & \textbf{Neg} & \textbf{Neu} & \textbf{Pos} & \textbf{Neg} & \textbf{Neu} & \textbf{Pos} & \textbf{Neg} & \textbf{Neu} & \textbf{Pos} & \textbf{Total} \\
\midrule
P1 & Pre-Escalation of Revolution (Jul 5--15)         &  2{,}281 &    853 &  1{,}576 &    489 &    183 &    338 &    488 &    182 &    337 &   6{,}727 \\
P2 & Crisis \& Blackout (Jul 16--Aug 4) &  9{,}180 &  4{,}700 &  9{,}698 &  1{,}967 &  1{,}007 &  2{,}078 &  1{,}968 &  1{,}008 &  2{,}078 &  33{,}684 \\
P3 & Post-Blackout (Aug 5--10)          & 14{,}153 &  7{,}482 & 10{,}704 &  3{,}033 &  1{,}603 &  2{,}294 &  3{,}032 &  1{,}603 &  2{,}293 &  46{,}197 \\
P4 & Post-Revolution (Aug 11--20)       & 23{,}664 & 10{,}532 & 14{,}479 &  5{,}071 &  2{,}257 &  3{,}103 &  5{,}070 &  2{,}257 &  3{,}103 &  69{,}536 \\
P5 & Flood Crisis (Aug 21--31)          & 17{,}104 &  5{,}682 &  7{,}493 &  3{,}665 &  1{,}218 &  1{,}606 &  3{,}665 &  1{,}217 &  1{,}605 &  43{,}255 \\
\midrule
\multicolumn{2}{l}{\textbf{Overall}}    & \textbf{66{,}382} & \textbf{29{,}249} & \textbf{43{,}950} & \textbf{14{,}225} & \textbf{6{,}268} & \textbf{9{,}419} & \textbf{14{,}223} & \textbf{6{,}267} & \textbf{9{,}416} & \textbf{199{,}399} \\
\bottomrule
\end{tabular}%
}
\end{table*}

\section{Dataset}
\label{sec:dataset}

We construct \dataset{}, a Bangla sentiment dataset based on public Facebook and YouTube discussions about the July--August 2024 Bangladesh uprising~\cite{RANA2026100341}. The dataset contains $\approx$200K annotated comments. Each comment is linked to its parent post, which allows us to evaluate sentiment classification in both comment-only and post+comment settings. Each comment is assigned one of three sentiment labels: \textsc{positive}, \textsc{neutral}, or \textsc{negative}. For each instance, we retain the comment text, parent post, timestamp, platform, and event phase.

\subsection{Data Collection}
\label{subsec:data_collection}

We collected public posts and comments from Facebook and YouTube. The posts were selected from verified news pages, public-interest groups, and high-engagement discussion threads related to the July--August 2024 uprising. We used the Apify platform\footnote{\url{https://apify.com}}, the Facebook Graph API, and the YouTube Data API v3 for data collection. The collection period covers July 5 to August 31, 2024. We divide this period into five phases: P1 Pre-Escalation (July 5--15), P2 Crisis and Blackout (July 16--August 4), P3 Post-Blackout (August 5--10), P4 Post-Revolution (August 11--20), and P5 Flood Crisis (August 21--31). These phases are used for temporal analysis and split construction. They are not used as sentiment labels. In total, we collected more than 2,000+ source posts and their comment threads. For each comment, we keep the corresponding parent post so that the dataset can support both comment-only and post+comment sentiment classification.

\subsection{Preprocessing}
\label{subsec:preprocessing}

We applied several preprocessing steps before annotation. First, we normalized the text using Unicode NFC normalization and removed irregular whitespace. URLs and user mentions were replaced with special tokens. Emojis were kept because they often carry sentiment in social media comments. We removed comments with fewer than three tokens, comments consisting mainly of non-Bangla text, and obvious spam entries. We also removed near-duplicate comments within the same thread using a Jaccard similarity threshold of 0.85. After preprocessing, the final dataset contains $\approx$200K comments.

\subsection{Annotation Guidelines}
\label{subsec:annotation_guidelines}

We prepared annotation guidelines for three-way sentiment classification. Annotators were asked to assign one label to each comment.

\noindent\textbf{Positive.}
A comment is labeled positive if it expresses support, approval, hope, praise, relief, or solidarity toward the event, actor, or situation discussed in the parent post.

\noindent\textbf{Neutral.}
A comment is labeled neutral if it is factual, unclear, mixed, question-like, or does not express a clear positive or negative sentiment.

\noindent\textbf{Negative.}
A comment is labeled negative if it expresses criticism, anger, frustration, sadness, fear, ridicule, sarcasm, blame, or disappointment. Annotators were instructed to use the parent post as context, but the label was assigned only to the comment. The guidelines included examples for difficult cases, including sarcasm, rhetorical questions, religious expressions, political references, code-switched profanity, and emoji-based sentiment. This follows standard practice in corpus annotation, where guidelines are refined through pilot annotation and disagreement analysis~\cite{pustejovsky-2012-role,sabou-etal-2014-corpus}.

\subsection{Manual Annotation}
\label{subsec:manual_annotation}

The annotation was carried out by 14 native Bangla-speaking undergraduate annotators from Bangladesh. All annotators were familiar with the July--August 2024 events and completed pilot training before the main annotation stage. A separate group of 10 senior validators supervised the process. Each comment was annotated independently by two annotators. If both annotators selected the same label, that label was accepted. If they disagreed, the comment was reviewed by a senior validator. Difficult cases were resolved through adjudication based on the annotation guidelines. The annotation process had five stages: pilot guideline development, annotator training and qualification, double annotation, senior adjudication, and blind audit. The full process took approximately eight months.

\subsection{Annotation Agreement}
\label{subsec:annotation_agreement}
We measured annotation reliability using Cohen's $\kappa$ and Krippendorff's $\alpha$. The annotation process achieved a pairwise Cohen's $\kappa$ of 0.73 and Krippendorff's $\alpha$ of 0.71, indicating substantial agreement. We also conducted a blind audit on a stratified subset of the dataset. The audit achieved 94.2\% exact-label agreement with the final gold labels. These results show that the annotation process was reliable despite the context-dependent nature of the comments.

\subsection{Dataset Split}
\label{subsec:data_split}

We split the dataset into training, validation, and test sets using a 70/15/15 ratio. The split was stratified by both sentiment label and event phase so that each partition preserves the overall class and temporal distribution. The training set contains 66,382 negative, 29,249 neutral, and 43,950 positive examples. The validation set contains 14,225 negative, 6,268 neutral, and 9,419 positive examples. The test set contains 14,223 negative, 6,267 neutral, and 9,416 positive examples. The full phase-wise distribution is shown in Table~\ref{tab:split_phase_sentiment}. Overall, the dataset contains 94,830 negative (47.6\%), 62,785 positive (31.5\%), and 41,784 neutral (21.0\%) comments.

\begin{table*}[t]
\centering
\caption{Zero-shot and few-shot performance (\%) of LLMs with expert prompting (Bangla-native) on UnRestSent200K. Best result per section is \textbf{bolded and underlined}.}
\label{tab:expert_bn}

\scriptsize

\resizebox{\textwidth}{!}{%
\begin{tabular}{lcccccccc}
\toprule

\multirow{2}{*}{\textbf{Model}}
& \multicolumn{4}{c}{\textbf{Zero-Shot}}
& \multicolumn{4}{c}{\textbf{Few-Shot}} \\

\cmidrule(lr){2-5}
\cmidrule(lr){6-9}

& \textbf{Acc}
& \textbf{F1}
& \textbf{MCC}
& \textbf{$\kappa$}
& \textbf{Acc}
& \textbf{F1}
& \textbf{MCC}
& \textbf{$\kappa$} \\

\midrule

\rowcolor{sectiongray}
\multicolumn{9}{l}{\textbf{Proprietary Models}} \\

GPT-4o-mini
& 71.0
& 69.0
& \underline{\textbf{50.0}}
& \underline{\textbf{47.0}}
& \underline{\textbf{76.0}}
& \underline{\textbf{75.0}}
& \underline{\textbf{58.0}}
& \underline{\textbf{55.0}} \\

gpt-4.1-mini
& \underline{\textbf{73.0}}
& \underline{\textbf{71.0}}
& 36.0
& 34.0
& 73.0
& 71.0
& 45.0
& 40.0 \\

gpt-4.1-nano
& 58.0
& 56.0
& 22.0
& 13.0
& 69.0
& 61.0
& 31.0
& 29.0 \\

\midrule

\rowcolor{sectiongray}
\multicolumn{9}{l}{\textbf{Open-Source Models ($\leq$3B)}} \\

\rowcolor{lightyellow}
gemma4-2b
& \underline{\textbf{61.0}}
& \underline{\textbf{59.0}}
& \underline{\textbf{40.0}}
& \underline{\textbf{38.0}}
& \underline{\textbf{64.0}}
& \underline{\textbf{63.0}}
& \underline{\textbf{45.0}}
& \underline{\textbf{43.0}} \\

\rowcolor{lightyellow}
gemma3-1b
& 43.0 & 37.0 & 18.0 & 15.0
& 52.0 & 47.0 & 26.0 & 22.0 \\

\rowcolor{lightyellow}
qwen2.5-3b
& 49.0 & 44.0 & 25.0 & 20.0
& 55.0 & 52.0 & 31.0 & 28.0 \\

\rowcolor{lightyellow}
qwen2.5-1.5b
& 46.0 & 36.0 & 20.0 & 11.0
& 49.0 & 40.0 & 23.0 & 15.0 \\

\rowcolor{lightyellow}
llama-3.2-3b
& 43.0 & 29.0 & 15.0 & 5.0
& 45.0 & 33.0 & 17.0 & 8.0 \\

\rowcolor{lightyellow}
qwen2.5-0.5b
& 38.0 & 22.0 & 0.2 & 0.0
& 39.0 & 23.0 & 2.0 & 0.4 \\

\midrule

\rowcolor{sectiongray}
\multicolumn{9}{l}{\textbf{Open-Source Models ($>$3B)}} \\

\rowcolor{lightblue}
gemma4-4b
& 52.0
& 49.0
& 16.0
& 15.0
& \underline{\textbf{66.2}}
& \underline{\textbf{65.0}}
& \underline{\textbf{48.7}}
& \underline{\textbf{46.3}} \\

\rowcolor{lightblue}
gemma3-4b
& 49.0 & 40.0 & 27.0 & 16.0
& 59.0 & 54.0 & 39.0 & 33.0 \\

\rowcolor{lightblue}
qwen3-4b
& \underline{\textbf{56.0}}
& \underline{\textbf{52.0}}
& \underline{\textbf{35.0}}
& \underline{\textbf{30.0}}
& 58.0
& 56.0
& \underline{\textbf{38.0}}
& \underline{\textbf{34.0}} \\

\rowcolor{lightblue}
qwen3-4b-instruct
& 56.0
& 52.0
& 35.0
& 30.0
& 59.0
& 57.0
& 23.0
& 14.0 \\

\rowcolor{lightblue}
qwen2.5-7b
& 47.0 & 45.0 & 23.0 & 21.0
& 61.0 & 60.0 & 42.0 & 39.0 \\

\rowcolor{lightblue}
llama-3.1-8b
& 53.0 & 52.0 & \underline{\textbf{37.0}} & 11.0
& 58.0 & 55.0 & 37.0 & 33.0 \\

\midrule
& \multicolumn{4}{c}{\textbf{ LoRA SFT}} \\
\cmidrule(lr){2-5}
\rowcolor{sectiongray}
\multicolumn{9}{l}{\textbf{Fine-Tuned Models}} \\

Gemma-4B
& \underline{\textbf{78.4}}
& \underline{\textbf{77.6}}
& \underline{\textbf{65.3}}
& \underline{\textbf{63.1}}
& --
& --
& --
& --
 \\

Llama-3.1-8B
& 76.8
& 75.9
& 62.4
& 60.2
& --
& --
& --
& --
 \\

\bottomrule
\end{tabular}
}
\end{table*}

\begin{table*}[t]
\centering
\caption{\textbf{Context ablation on \dataset{}.} Fine-tuned encoder performance when the model receives only the \emph{comment} versus the \emph{parent post concatenated with the comment} via \texttt{[SEP]}. Corpus, splits, optimiser, schedule, and hyperparameters are held fixed; the only variable is the input. Adding the parent post yields a consistent $+7$--$11$pp accuracy gain across every architecture. Best per column \underline{\textbf{bolded and underlined}}.}
\label{tab:model_comparison_combined}

\setlength{\tabcolsep}{5pt}
\renewcommand{\arraystretch}{1.2}

\begin{tabular}{
l
r
r
r
r
r
r
r
r
}
\toprule
& \multicolumn{4}{c}{\textbf{Comment-only input}} & \multicolumn{4}{c}{\textbf{Post\,+\,Comment input}} \\
& \multicolumn{4}{c}{\scriptsize\texttt{[comment]}} & \multicolumn{4}{c}{\scriptsize\texttt{[post]\,[SEP]\,[comment]}} \\
\cmidrule(lr){2-5} \cmidrule(lr){6-9}
\textbf{Model} &
{\textbf{Acc.\,(\%)}} &
{\textbf{Prec.}} &
{\textbf{Rec.}} &
{\textbf{F1}} &
{\textbf{Acc.\,(\%)}} &
{\textbf{Prec.}} &
{\textbf{Rec.}} &
{\textbf{F1}} \\
\midrule
BanglaBERT        & \multicolumn{1}{c}{\underline{\bfseries 64.19}} & 0.6195 & 0.6158 & 0.6170 & \multicolumn{1}{c}{\underline{\bfseries 74.72}} & 0.7245 & 0.7116 & 0.7167 \\
mBERT             & 61.06 & 0.5869 & 0.5881 & 0.5874 & 69.90 & 0.6705 & 0.6790 & 0.6731 \\
BanglaElectra     & 62.02 & 0.6193 & 0.6202 & 0.6159 & 68.98 & 0.6637 & 0.6705 & 0.6649 \\
XLM-RoBERTa-Base  & 63.41 & 0.6287 & 0.6312 & 0.6299 & 72.96 & 0.7289 & 0.7296 & 0.7012 \\
\bottomrule
\end{tabular}
\end{table*}

\begin{table}[t]
    \centering
\caption{Temporal generalization performance of supervised fine-tuned (SFT) transformer models under distribution shift on UnRestSent200K. Models are trained on comments collected between July 5 and August 5, 2024, and evaluated on two chronologically non-overlapping future test partitions: Test-A (August 6–20) and Test-B (August 21–31). Results highlight the degradation of model performance under evolving crisis discourse and temporal non-stationarity.}
    \label{tab:time_shift_performance}
    \scriptsize
    \setlength{\tabcolsep}{2.5pt}
    \renewcommand{\arraystretch}{1.05}

    \resizebox{\columnwidth}{!}{%
    \begin{tabular}{@{}
        l
        c
        r
        r
        r
        r
    @{} }
    \toprule
    \textbf{Model} &
    \textbf{Test} &
    {\textbf{Acc. (\%)}} &
    {\textbf{Prec.}} &
    {\textbf{Rec.}} &
    {\textbf{F1 Score}} \\
    \midrule
     \multirow{2}{*}{BanglaBERT}
        & A & 53.70 & 0.5204 & 0.5236 & 0.5217 \\
        & B & \multicolumn{1}{c}{\underline{\bfseries 54.83}} & 0.5027 & 0.5222 & 0.5064  \\
    \midrule

    \multirow{2}{*}{BanglaElectra}
        & A & 40.11 & 0.4195 & 0.4275 & 0.4006 \\
        & B & 41.18 & 0.4164 & 0.4343 & 0.3942  \\
    \midrule

    \multirow{2}{*}{mBERT}
        & A & 40.96 & 0.4234 & 0.4380 & 0.4111  \\
        & B & 41.24 & 0.4191 & 0.4354 & 0.3948  \\
    \midrule

    \multirow{2}{*}{XLM-RoBERTa}
        & A & 47.35 & 0.4708 & 0.4784 & 0.4695 \\
        & B & 47.28 & 0.4480 & 0.4664 & 0.4436  \\
    \bottomrule

    \end{tabular}%
    }
\end{table}
\section{Experiments and Analysis}
\label{sec:experiments}

The July--August 2024 Bangladesh uprising was not a static event~\cite{RANA2026100341}. Public discussion changed as the crisis moved from early protest activity to internet blackout, regime collapse, political transition, and later the flood crisis. This makes \dataset{} useful for studying more than standard sentiment classification. It allows us to ask whether sentiment models can handle three problems that often appear in real crisis discourse: context dependence, temporal change, and abrupt shifts in public reaction. We organize the experiments around these problems. First, we evaluate recent LLMs to see how far prompting alone can go on Bangla crisis sentiment. Second, we test whether supervised adaptation with mid-scale LLMs improves over prompting. Third, we measure the effect of adding the parent post as context. Finally, we evaluate temporal robustness and examine whether sentiment changes gradually or through phase-level shifts. We evaluate four fine-tuned encoder models: BanglaBERT~\cite{Bhattacharjee2022BanglaBERT}, BanglaElectra, mBERT~\cite{devlin2019bert}, and XLM RoBERTa~\cite{conneau2020unsupervised}. We also evaluate 15 LLMs under zero-shot and few-shot prompting. Across experiments, we report accuracy, macro-F1, and chance-corrected metrics such as Cohen's $\kappa$ and MCC where applicable.

\subsection{Prompted LLMs}
\label{subsec:llm_eval}

\noindent\textbf{Motivation.}
We first ask whether recent LLMs can classify sentiment in \dataset{} without task-specific training. This is an important baseline because LLMs are often used directly on new events, especially when labeled data is limited.

\noindent\textbf{Models and prompting setup.}
We evaluate open-source models from the Gemma, Qwen, and LLaMA families, along with proprietary models including GPT-4o-mini, GPT-4.1-mini, and GPT-4.1-nano. Each model is tested with zero-shot and few-shot prompts written in Bangla. In the few-shot setting, the prompt includes seven examples covering positive, negative, neutral, sarcastic, and politically implicit comments. The full prompt templates are shown in Appendix~\ref{appendix:sample_prompt_result}.

\noindent\textbf{Input format.}
Each test input is given in the same format: \texttt{[post]\,[SEP]\,[comment]}. This is the same input format used in the post+comment encoder setting, making the LLM and encoder results comparable.

\noindent\textbf{Main results.}
Table~\ref{tab:expert_bn} shows that the strongest prompted model is GPT-4o-mini in the few-shot setting. It reaches 76.0\% accuracy, 75.0 macro-F1, and $\kappa=0.55$. This is slightly above the strongest post+comment encoder, BanglaBERT, which reaches 74.72\% accuracy and 0.7167 macro-F1. However, the improvement is small. This suggests that strong LLMs can benefit from the provided context, but prompting alone does not fully solve the task.

\noindent\textbf{Effect of few-shot examples.}
Few-shot examples help the stronger models more than the smaller ones. For GPT-4o-mini, $\kappa$ improves from 0.47 to 0.55. For GPT-4.1-mini, it improves from 0.34 to 0.40. In contrast, several smaller open-source models show little improvement from few-shot examples. This suggests that examples are useful only when the model already has enough Bangla language ability and event-level reasoning capacity to use them.

\noindent\textbf{Error patterns.}
A manual inspection of errors shows three recurring patterns. First, models often miss sarcasm, especially when a comment appears positive on the surface but is negative in context. Second, neutral comments are often pushed toward negative labels. Third, politically charged comments become harder when their interpretation depends on the phase of the uprising. These errors are also seen in the encoder models, suggesting that they are properties of the task rather than failures of one model family.

\subsection{Supervised Adaptation with Mid-Scale LLMs}
\label{subsec:sft_gemma}

Prompting provides a useful starting point, but \dataset{} also allows us to test whether in-domain supervised training improves performance. We therefore fine-tune two mid-scale LLMs, Gemma-4B and LLaMA-3.1-8B, using LoRA. Both models are trained with $r{=}64$, $\alpha{=}32$, and learning rate $3{\times}10^{-4}$. The fine-tuned models outperform all prompted models in Table~\ref{tab:expert_bn}. Gemma-4B achieves 78.4\% accuracy, 77.6 macro-F1, MCC 65.3, and $\kappa$ 63.1. LLaMA-3.1-8B achieves 76.8\% accuracy and 75.9 macro-F1. Gemma-4B also improves over GPT-4o-mini by 2.4 percentage points in accuracy and over grounded BanglaBERT by 3.7 points. These results show that supervised adaptation is still important for low-resource crisis sentiment analysis. The smaller Gemma-4B model also performs better than the larger LLaMA-3.1-8B model, suggesting that the choice of backbone and fine-tuning recipe matters more than parameter count alone. At the same time, the best model remains below 80\% accuracy, leaving room for future work on crisis-aware and context-aware sentiment models.

\subsection{Qualitative Effect of Fine-Tuning}
\label{sec:sft_qualitative}

Table~\ref{tab:sft_vs_zeroshot} illustrates the three failure patterns flagged in \S\ref{subsec:llm_eval} sarcasm, neutral/positive collapse to negative, and politically charged surface cues and shows the same \texttt{(post, comment)} pairs being recovered after LoRA-SFT on Gemma-4B. The pattern is qualitative: SFT exposure to the gold PiLA labels lets the model pick up domain-specific pragmatic cues that the zero-shot model misses despite seeing identical input.

\definecolor{headerblue}{HTML}{EAF1F8}
\definecolor{analysisblue}{HTML}{F4F7FA}
\definecolor{translationgray}{HTML}{555D66}
\definecolor{correctgreen}{HTML}{1B7F3A}
\definecolor{errorred}{HTML}{C62828}

\newcommand{\cmark}{\textcolor{correctgreen}{\ding{51}}}
\newcommand{\xmark}{\textcolor{errorred}{\ding{55}}}
\newcommand{\pred}[2]{\textbf{#1}\,#2}
\newcommand{\trans}[1]{%
  {\color{translationgray}\footnotesize ``#1''}%
}

\begin{table*}[t]
\centering
\small
\setlength{\tabcolsep}{5pt}
\renewcommand{\arraystretch}{1.15}

\caption{
\textbf{Qualitative comparison of zero-shot and supervised predictions.}
Both settings use Gemma-4B; LoRA SFT denotes the same backbone fine-tuned
on \dataset{} ($r{=}64$, $\alpha{=}32$, learning rate $3{\times}10^{-4}$).
\cmark{} and \xmark{} indicate agreement and disagreement with the PiLA
gold label, respectively. Examples represent the three failure modes
discussed in \S\ref{subsec:llm_eval}.
}
\label{tab:sft_vs_zeroshot}

\begin{tabularx}{\textwidth}{
    @{}
    >{\centering\arraybackslash}p{0.38cm}
    >{\raggedright\arraybackslash}X
    >{\centering\arraybackslash}p{1.15cm}
    >{\centering\arraybackslash}p{1.50cm}
    >{\centering\arraybackslash}p{1.50cm}
    @{}
}
\toprule
\rowcolor{headerblue}
\textbf{\#}
&
\textbf{Post--Comment Pair}
&
\textbf{Gold}
&
\textbf{Zero-shot}
&
\textbf{LoRA SFT}
\\
\midrule

\textbf{1}
&
\textbf{Post}\quad
\bn{আমার বাসায় কাজ করা পিয়ন এখন ৪০০ কোটি টাকার মালিক: প্রধানমন্ত্রী}

\trans{The peon who worked at my house now owns 4 billion taka:
Prime Minister.}

\smallskip
\textbf{Comment}\quad
\bn{আহ কি গর্বের কথা!}\enspace
\trans{Ah, what a matter of pride!}
&
\textbf{Negative}
&
\pred{Positive}{\xmark}
&
\pred{Negative}{\cmark}
\\[-1pt]

\rowcolor{analysisblue}
\multicolumn{5}{
    @{}>{\raggedright\arraybackslash}p{\textwidth}@{}
}{
\footnotesize\color{translationgray}
\textbf{\textit{Sarcasm.}}
The ostensibly positive expression ``pride'' becomes negative under the
corruption-related headline. Zero-shot follows the literal wording,
whereas LoRA SFT captures the intended sarcasm.
}
\\
\addlinespace[4pt]

\textbf{2}
&
\textbf{Post}\quad
\bn{দেশে ইন্টারনেট সেবায় ধীরগতি, ফেসবুক–মেসেঞ্জারে প্রবেশেও বিঘ্ন}

\trans{Internet service is slow nationwide; access to Facebook and
Messenger is also disrupted.}

\smallskip
\textbf{Comment}\quad
\bn{এটা হাসিনার কাজ}\enspace
\trans{This is Hasina's doing.}
&
\textbf{Negative}
&
\pred{Neutral}{\xmark}
&
\pred{Negative}{\cmark}
\\[-1pt]

\rowcolor{analysisblue}
\multicolumn{5}{
    @{}>{\raggedright\arraybackslash}p{\textwidth}@{}
}{
\footnotesize\color{translationgray}
\textbf{\textit{Pragmatic negativity.}}
Zero-shot treats the comment as a neutral factual attribution, whereas
LoRA SFT recognizes its accusatory and negative pragmatic force.
}
\\
\addlinespace[4pt]

\textbf{3}
&
\textbf{Post}\quad
\bn{বন্যায় এখন পর্যন্ত ১৩ জনের মৃত্যু, ক্ষতিগ্রস্ত ৪৫ লাখ মানুষ}

\trans{13 dead in floods so far; 4.5 million affected.}

\smallskip
\textbf{Comment}\quad
\bn{আল্লাহ আপনি সবাইকে হেফাজত করেন।}\enspace
\trans{May Allah protect everyone.}
&
\textbf{Positive}
&
\pred{Negative}{\xmark}
&
\pred{Positive}{\cmark}
\\[-1pt]

\rowcolor{analysisblue}
\multicolumn{5}{
    @{}>{\raggedright\arraybackslash}p{\textwidth}@{}
}{
\footnotesize\color{translationgray}
\textbf{\textit{Surface-cue confusion.}}
Disaster-related terms in the post mislead the zero-shot model, while
LoRA SFT correctly identifies the prayer and expression of solidarity
as positive.
}
\\

\bottomrule
\end{tabularx}
\end{table*}

\subsection{Effect of Parent-Post Context}
\label{subsec:context_impact}

Many comments in \dataset{} are difficult to interpret without their parent post, as shown in Table~\ref{tab:sft_vs_zeroshot}. This is common in social media discussions during political events. A short comment may look factual or neutral by itself, but become clearly positive or negative once the surrounding news post is known. For example, the comment \emph{``We know who is doing this.''} is hard to label in isolation. When paired with a post about a family being forced from their home, the same comment becomes a negative reaction. This shows why the parent post is not extra information; in many cases, it is part of the meaning of the comment. We test this directly using a paired ablation. In the comment-only setting, the encoder receives only the comment. In the post+comment setting, the encoder receives the parent post and comment separated by \texttt{[SEP]}. All other training settings are kept fixed. The results in Table~\ref{tab:model_comparison_combined} show consistent gains from context. BanglaBERT improves from 64.19\% to 74.72\% accuracy. XLM-RoBERTa improves by 9.55 points, mBERT by 8.84 points, and BanglaElectra by 6.96 points. The gain appears across all encoder families. This result has a direct implication for future dataset design. In crisis discourse, comment-only sentiment classification can remove information needed for the correct label. For this reason, \dataset{} supports both comment-only and post+comment evaluation, allowing future work to measure how well models use discourse context.

\subsection{Temporal Robustness}
\label{subsec:temporal_robustness}

We next evaluate whether models trained on earlier phases of the uprising generalize to later phases. This setting reflects a realistic use case: during an unfolding crisis, models may be trained on available data and then applied as the event continues to change. We train the encoder models on comments from July 5 to August 5, 2024. We then test them on two later windows: Test-A, covering August 6--20, and Test-B, covering August 21--31. These windows do not overlap with the training period. The results in Table~\ref{tab:time_shift_performance} show a large performance drop. BanglaBERT remains the best model, reaching 54.83\% accuracy on Test-B, but this is far below its in-distribution post+comment accuracy of 74.72\%. Other encoders also degrade under the time-based split. Overall, the models lose between 19 and 28 macro-F1 points compared with their in-distribution results. This drop is important because the language, platforms, and broad event remain the same. What changes is the stage of the crisis. After the blackout and regime collapse, public discussion shifts toward new actors, new concerns, and new emotional frames. Appendix~\ref{app:topic_results} further shows measurable lexical change across phases using Jensen--Shannon divergence. These results show why random splits are not enough for crisis sentiment analysis. A model that performs well on a random test set may still fail when the event enters a new phase. \dataset{} therefore provides a useful setting for studying temporal robustness in low-resource NLP.

\subsection{Sentiment Change Across Phases}
\label{subsec:sentiment_dynamics}

The temporal experiment shows that performance drops over time. We then ask whether this change is gradual or concentrated around specific moments in the uprising. To measure this, we compute Cohen's $d$ for all pairwise comparisons between the five event phases. The full results are shown in Table~\ref{tab:period_comparisons}. Seven of the ten comparisons show negligible effect sizes. The strongest changes are centered around the Post-Blackout phase. The comparison between Crisis and Blackout and Post-Blackout has a medium effect size ($d=-0.556$), and the comparison between Post-Blackout and Flood Crisis also has a medium effect size ($d=0.567$). This suggests that sentiment did not change smoothly across the full period. Instead, the largest shift occurred around the moment when internet access returned, the regime collapsed, and people began reacting to a new political situation. Sentiment then shifted again as the flood crisis became the dominant concern. For future research, this makes \dataset{} useful for studying change-point-aware sentiment models. Models for crisis monitoring may need to detect when the discourse has entered a new phase, rather than assuming that sentiment changes slowly over time.

\begin{table}[t]
\centering
\scriptsize
\caption{Pairwise sentiment-shift comparisons between the five event phases of the 2024 Bangladesh uprising using Cohen's $d$ effect size. Larger absolute values indicate stronger changes in sentiment distribution between phases, with the strongest shifts observed around the immediate post-blackout period.}
\label{tab:period_comparisons}

\setlength{\tabcolsep}{3pt}

\begin{tabular}{llcl}
\toprule
\textbf{Period 1} & \textbf{Period 2} & \textbf{$d$} & \textbf{Effect} \\
\midrule

Pre-Escalation & Crisis/Blackout & 0.138 & Negligible \\

Pre-Escalation & Post-Blackout & $-0.411$ & Small \\

Pre-Escalation & Post-Revolution & $-0.034$ & Negligible \\

Pre-Escalation & Flood Crisis & 0.144 & Negligible \\

Crisis/Blackout & Post-Blackout & $-0.556$ & Medium \\

Crisis/Blackout & Post-Revolution & $-0.174$ & Negligible \\

Crisis/Blackout & Flood Crisis & 0.003 & Negligible \\

Post-Blackout & Post-Revolution & 0.381 & Small \\

Post-Blackout & Flood Crisis & 0.567 & Medium \\

Post-Revolution & Flood Crisis & 0.178 & Negligible \\

\bottomrule
\end{tabular}

\vspace{-0.2cm}
\end{table}

\subsection{Findings}
\label{subsec:findings}

Our experiments yield five main findings.

\noindent\textbf{Prompting alone does not resolve crisis sentiment.}
Few-shot GPT-4o-mini achieves the strongest prompted result at 76.0\% accuracy, only slightly exceeding grounded BanglaBERT at 74.72\%. Its remaining errors in sarcasm, implicit attribution, and context-dependent language show that strong prompting cannot fully address the task.

\noindent\textbf{Supervised adaptation produces the best performance.}
LoRA-tuned Gemma-4B achieves the highest overall accuracy of 78.4\%, outperforming both prompted LLMs and fine-tuned encoders. As illustrated in Table~\ref{tab:sft_vs_zeroshot}, in-domain supervision helps recover pragmatic meanings that the zero-shot model misses, including sarcastic criticism, political accusation, and expressions of solidarity.

\noindent\textbf{Parent-post context is essential.}
Including the parent post improves all encoders by approximately 7--11 accuracy points. This consistent gain shows that the post is often part of the sentiment signal rather than optional background information.

\noindent\textbf{Temporal shift substantially reduces performance.}
Models trained on earlier phases lose approximately 19--28 macro-F1 points when evaluated on later phases. Thus, random splits can overestimate real-world reliability by mixing examples from different stages of an evolving crisis.

\noindent\textbf{Sentiment changes around major events.}
Most phase comparisons show negligible differences, but medium shifts occur around the Post-Blackout period ($d=-0.556$) and the later Flood Crisis ($d=0.567$). This suggests that crisis sentiment changes through event-linked transitions rather than a smooth temporal trend.

\noindent
Together, these findings show that reliable crisis sentiment analysis requires both discourse context and temporal adaptation. \dataset{} supports the study of these challenges through parent-linked comments and event-aligned evaluation phases.

\section{Discussion}
\label{sec:discussion}

Our results show that crisis sentiment analysis depends strongly on both context and time. Models perform much better when the parent post is included, with consistent gains of 7--11 percentage points across encoder models. At the same time, performance drops substantially when models are trained on earlier phases of the uprising and tested on later phases. This shows that standard random splits can overestimate model performance in an unfolding crisis. The results also show that strong LLMs do not remove these challenges. GPT-4o-mini performs well under few-shot prompting, but it only slightly improves over the best post+comment encoder and remains below the best fine-tuned LLM. Its errors also overlap with those of smaller models, especially on sarcasm, implicit political references, and comments whose meaning depends on the event phase. These findings suggest that future sentiment benchmarks, especially for crisis settings, should include temporal splits and discourse context. For low-resource languages such as Bangla, this is particularly important because models may appear reliable under standard evaluation while failing when the public conversation changes. \dataset{} provides a setting for studying these issues in a controlled way.

\section{Conclusion}
\label{sec:conclusion}

We introduce \dataset{}, a Bangla crisis sentiment dataset containing $\approx$200K comments from Facebook and YouTube discussions around the July--August 2024 Bangladesh uprising. Each comment is linked to its parent post and assigned one of three sentiment labels: \textsc{positive}, \textsc{neutral}, or \textsc{negative}. The dataset is annotated through a fully human process under the PiLA framework. Experiments with fine-tuned encoders, prompted LLMs, and LoRA-tuned LLMs show three main results. First, parent-post context consistently improves sentiment classification. Second, temporal shift across phases of the uprising leads to large performance drops. Third, supervised adaptation gives the strongest results, but the benchmark remains challenging. Overall, \dataset{} supports research on context-aware and temporally robust sentiment analysis in low-resource crisis discourse.

\newpage

\textbf{Limitations}
\label{sec:limitations}

\dataset{} has several limitations. First, it covers only Facebook and YouTube. These platforms were important sources of public discussion during the uprising, but they do not represent all Bangla speakers or all forms of political communication. Other sources, such as X/Twitter, Telegram, private messaging, and offline discourse, are not included. Second, the dataset uses three sentiment labels. This makes large-scale annotation reliable, but it does not capture finer distinctions such as anger, fear, hope, grief, or stance. Future work can extend the label schema to include emotion and stance labels. Third, the dataset is based on one event in Bangladesh. The results should therefore not be treated as universal claims about all crises or all low-resource languages. Cross-event and cross-language studies are needed to test how well the findings generalize. Finally, the models evaluated in this paper were not specifically designed for temporal adaptation or change-point detection. Developing models that can use context and adapt to new crisis phases is an important direction for future work.

\textbf{Reproducibility Statement}
\label{appendix:reproducibility}

To support reproducibility, the release package will include the processed and de-identified \dataset{} corpus; fixed training, validation, and test splits; parent-post context; timestamps; platform and event-phase metadata; annotation guidelines; prompt templates; preprocessing and evaluation scripts; and the configurations used for the encoder and LoRA experiments. Dataset statistics and phase boundaries are reported in Section~\ref{sec:dataset}, while detailed preprocessing, annotation, temporal-analysis, topic-modeling, and prompting procedures are provided in the remainder of this appendix.

\textbf{AI Usage Disclosure}
Large language models were used as experimental systems to generate benchmark predictions for \dataset{} under zero-shot and few-shot prompting and supervised fine-tuning. AI-assisted tools were also used for limited non-experimental support, including language editing, figure ideation, and visualization assistance. All AI-assisted material was manually reviewed and verified by the authors, who retain full responsibility for the paper's content, results, and interpretations.

\bibliography{ref}

\appendix
\section{Appendix}
\subsection{Reproducibility Statement}
\label{appendix:reproducibility}

To support reproducibility, the release package will include the processed and de-identified \dataset{} corpus; fixed training, validation, and test splits; parent-post context; timestamps; platform and event-phase metadata; annotation guidelines; prompt templates; preprocessing and evaluation scripts; and the configurations used for the encoder and LoRA experiments. Dataset statistics and phase boundaries are reported in Section~\ref{sec:dataset}, while detailed preprocessing, annotation, temporal-analysis, topic-modeling, and prompting procedures are provided in the remainder of this appendix.

\subsection{AI Usage Disclosure}
Large language models were used as experimental systems to generate benchmark predictions for \dataset{} under zero-shot and few-shot prompting and supervised fine-tuning. AI-assisted tools were also used for limited non-experimental support, including language editing, figure ideation, and visualization assistance. All AI-assisted material was manually reviewed and verified by the authors, who retain full responsibility for the paper's content, results, and interpretations.

\subsection{Preprocessing Details}
\label{appendix:preprocessing}

This section describes the preprocessing steps used before annotation, as summarized in Section~\ref{subsec:preprocessing}.

\subsubsection{Normalization}
\begin{itemize}[leftmargin=*,itemsep=2pt]
    \item \textbf{Unicode normalization:} Text is normalized to NFC form to reduce inconsistent Bangla character encodings.
    \item \textbf{Whitespace cleanup:} Extra spaces, tabs, and irregular line breaks are removed.
    \item \textbf{Entity masking:} URLs and user mentions are replaced with \texttt{[URL]} and \texttt{[USER]}.
    \item \textbf{Emoji retention:} Emojis are kept because they often express sentiment in social media comments.
\end{itemize}

\subsubsection{Filtering}
We remove comments with fewer than three tokens after normalization. We also remove comments that consist mainly of non-Bangla text or obvious spam. To keep the released corpus primarily Bangla, comments with substantial English or Bangla--English transliteration are excluded. Figure~\ref{fig:sentence_length} reports the resulting sentiment distribution across word-length bins of the retained corpus: short comments ($\le$10 words) skew positive, while neutral expression increases with comment length and negative sentiment stays relatively flat across lengths.

\subsubsection{Deduplication}
Near-duplicate comments within the same thread are removed using a Jaccard similarity threshold of 0.85. Duplicates across different posts are retained when they appear as part of separate public discussions.

\subsubsection{Post-Level Framing Labels}
In addition to comment-level sentiment labels, each source post is assigned a coarse framing label:
\begin{equation}
\mathcal{Y}^{(p)}=\{\textsc{Hope},\textsc{Despair},\textsc{Outrage}\}
\end{equation}

\noindent These labels describe the dominant framing of the parent post. \textsc{Hope} captures optimistic or change-oriented framing, \textsc{Despair} captures grief or uncertainty, and \textsc{Outrage} captures anger, blame, or calls for accountability. These post-level labels are used as annotation context and metadata; the main prediction task remains comment-level sentiment classification.

\subsection{PiLA: Annotation Protocol Details}
\label{appendix:annotation}

This appendix provides additional details about the \textbf{Pilot Label Annotation (PiLA)} framework shown in Figure~\ref{fig:annotation_overview}. PiLA is a fully human annotation process. No labels are assigned by automated systems. The full annotation effort took approximately eight months and involved 14 primary annotators and 10 senior validators.

\begin{table}[t]
    \centering
    \caption{Table showing topic modeling statistics for each period. $C_v$ denotes the topic coherence score, computed using normalized pointwise mutual information (NPMI) and ranging from $0$ to $1$, where higher values indicate greater semantic coherence. \textit{Eff.} represents the effective number of topics, calculated as $\exp(H)$, where $H = -\sum_i p_i \log p_i$ is the Shannon entropy of the topic distribution. This quantity reflects the number of equally prominent topics that would yield the same level of topic diversity.}
    \label{tab:period_topics}
    \resizebox{\columnwidth}{!}{%
    \begin{tabular}{lrccc}
    \toprule
    \textbf{Period} & \textbf{Docs} & \textbf{Topics} & \textbf{C$_v$} & \textbf{Eff.} \\
    \midrule
      P1: Pre-Escalation      & 6{,}727   & 12 & 0.405 & 1.7 \\
    P2: Crisis \& Blackout  & 33{,}684  & 29 & \textbf{0.781} & 1.8 \\
    P3: Post-Blackout       & 46{,}197  & 18 & 0.384 & 1.9 \\
    P4: Post-Revolution     & 69{,}536  & 25 & 0.420 & 2.0 \\
    P5: Flood Crisis        & 43{,}255  & 25 & 0.422 & 1.9 \\
    \midrule
    \textbf{Global}         & 199{,}399 & 29 & 0.375 & --- \\
    \bottomrule
    \end{tabular}%
    }
    \vspace{-0.1cm}
\end{table}
\subsubsection{Annotator and Validator Pool}
\label{app:pila_pool}

The primary annotation team consists of 14 undergraduate annotators from Bangladesh. All annotators are native Bangla speakers and were familiar with the July--August 2024 events through Bangla news and social media. The validator team consists of 10 senior annotators who did not overlap with the primary annotation pool. All participants provided informed consent, and the study protocol was reviewed by the host institution.

\subsubsection{Guideline Document}
\label{app:pila_guidelines}

The annotation guideline defines the three sentiment labels: \textsc{positive}, \textsc{neutral}, and \textsc{negative}. It also provides rules and examples for common ambiguous cases, including sarcasm, rhetorical questions, religious expressions, political references, code-switched profanity, and emoji-based sentiment. Annotators were instructed to use the parent post as context, but not to copy the post's sentiment into the comment label. The label was assigned to the comment only. The guideline was refined during the pilot stage and then fixed before the main annotation stage.

\subsubsection{Qualification and Quality Checks}
\label{app:pila_qualification}

Before the main annotation stage, annotators completed a qualification round. Only annotators who reached Cohen's $\kappa \geq 0.65$ were admitted to the main task. All 14 annotators met this threshold, with scores ranging from 0.66 to 0.81. During annotation, gold-check examples were inserted at a 2\% rate. Annotators whose agreement dropped below the required threshold were paused, retrained on the relevant ambiguity cases, and re-qualified before continuing. Labels from affected batches were reviewed and re-annotated where necessary.

\subsubsection{Adjudication}
\label{app:pila_logs}

Each comment was independently labeled by two annotators. If both annotators agreed, their shared label was accepted. If they disagreed, the comment was reviewed by a senior validator. Difficult cases were escalated to a small validator panel. For adjudicated comments, we record the two initial labels, annotator confidence scores, validator decision, and final gold label. These logs allow later analysis of disagreement patterns and label reliability.

\subsection{Quality Metrics Details}
\label{appendix:quality}

\subsubsection{Pairwise Inter-Annotator Agreement}

We report Cohen's $\kappa$ for agreement between the two primary annotators assigned to each comment. The overall pairwise score is $\kappa = 0.73$, indicating substantial agreement. We compute:
\begin{equation}
\kappa = \frac{p_o - p_e}{1 - p_e},
\end{equation}
where $p_o$ is observed agreement and $p_e$ is expected agreement under the annotators' marginal label distributions. Per-phase $\kappa$ ranges from 0.69 to 0.78.

\subsubsection{Multi-Rater Reliability}

We also report Krippendorff's $\alpha$ to measure reliability across the full annotator pool. The overall score is $\alpha = 0.71$, showing that the labels are consistent beyond individual annotator pairs.

\subsubsection{Annotator Confidence}

Annotators provided a confidence score in $[0,1]$ for each label. The mean confidence score is $\bar{c}=0.82$ with standard deviation 0.14. In total, 87.3\% of labels have confidence scores of at least 0.7. These scores are included as metadata so that users can filter examples by confidence if needed.

\subsubsection{Pre-Adjudication Agreement}

The two primary annotators agreed directly on 78.4\% of comments. The remaining 21.6\% were sent to senior validators for adjudication. Among these, 3.1\% required review by a three-validator panel.

\subsubsection{Independent Blind Audit}

As a final quality check, two senior validators who were not involved in adjudication re-labeled a stratified random sample of 2,000 comments, with 400 comments from each phase. They were blind to the final gold labels. The audit achieved 94.2\% exact-label agreement and 96.8\% agreement within one step on the sentiment scale. Most disagreements involved rhetorical questions and emoji-mediated sentiment, which were also identified as difficult cases during guideline development.

\subsection{Effect of Comment Length}
\label{appendix:comment_length}

\begin{figure}[t]
    \centering
    \includegraphics[width=\columnwidth]{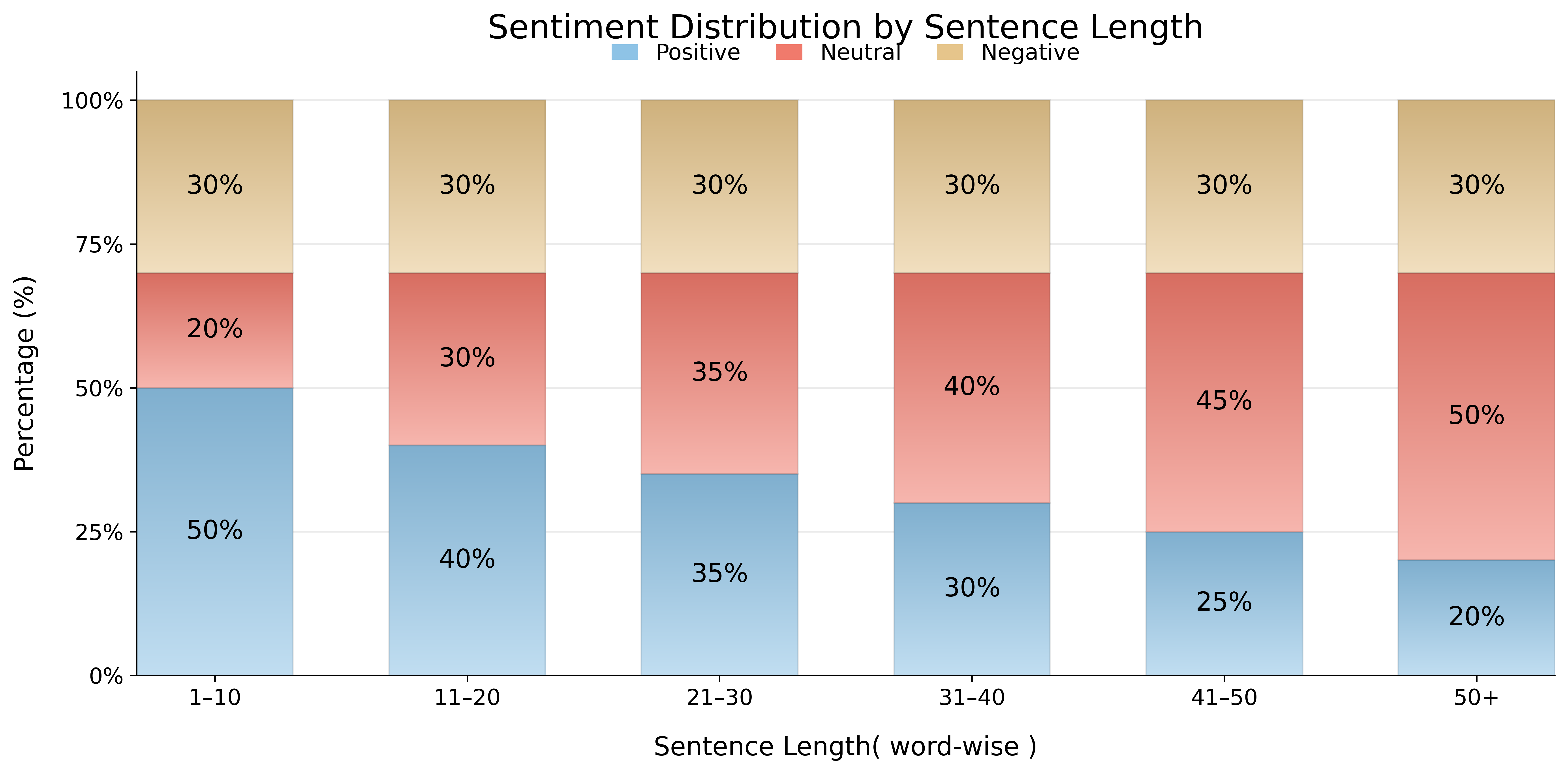}
    \caption{Sentiment distribution across sentence length intervals (word-wise). Short sentences (1--10 words) are dominated by positive sentiment, while the proportion of neutral sentiment increases with sentence length. Negative sentiment remains relatively stable across all length ranges.}
    \label{fig:sentence_length}
    \vspace{-0.3cm}
\end{figure}

We analyze sentiment distribution across word-level comment length intervals to examine whether affective polarity varies with comment verbosity. As shown in Figure~\ref{fig:sentence_length}, shorter comments are more frequently positive, with the positive share decreasing from 50\% in 1--10 word comments to 20\% in comments longer than 50 words. In contrast, neutral sentiment increases steadily with length, suggesting that longer comments tend to provide explanation, context, or factual discussion rather than direct affective reactions. Negative sentiment remains comparatively stable across all length ranges. This indicates that comment length is associated with sentiment composition, and that models may need to account for verbosity-driven shifts in crisis discourse.

\subsection{Quantifying Temporal Non-Stationarity}
\label{appendix:temporal_nonstationarity}

\begin{figure}[H]
    \centering
    \includegraphics[width=\columnwidth]{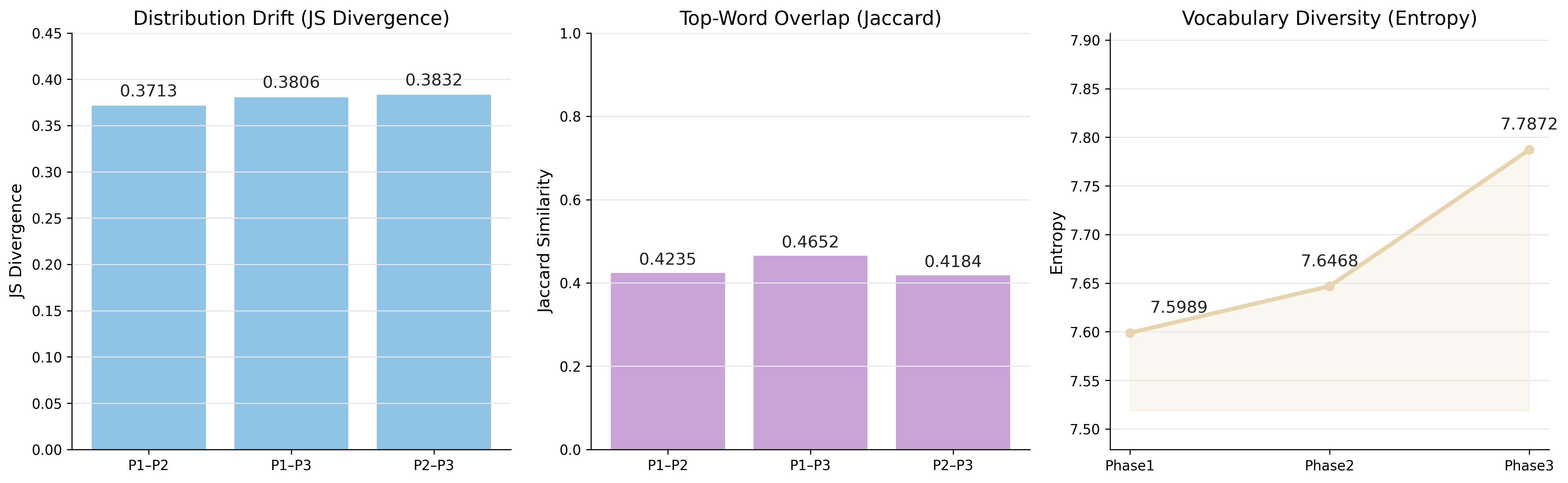}
\caption{Temporal drift across three chronological phases of Bangla text. Divergence, vocabulary turnover, and rising entropy together reveal substantial non-stationarity and increasing lexical diversity over time.}
    \label{fig:temporal_data_drift}
    \vspace{-0.3cm}
\end{figure}

We measure lexical change across event phases using three signals: Jensen--Shannon (JS) divergence over unigram distributions, Jaccard similarity of top words, and Shannon entropy. The results are shown in Figure~\ref{fig:temporal_data_drift}.

\paragraph{JS divergence.}
JS divergence values range from 0.3713 to 0.3832 across phase pairs, showing that word distributions change over time.

\paragraph{Jaccard similarity.}
Top-word Jaccard similarity ranges from 0.4235 to 0.4652. This means that less than half of the most frequent terms are shared across phase pairs, indicating vocabulary turnover as the event develops.

\paragraph{Entropy.}
Shannon entropy increases from 7.5989 to 7.7872, suggesting that the discourse becomes more lexically diverse in later phases. Together, these results show that the corpus changes measurably across event phases. This supports the temporal-shift results in Table~\ref{tab:time_shift_performance}: performance drops are associated with changes in the data distribution, not only with model limitations. The corresponding sentiment-level shift, quantified using Cohen's $d$ on the polarity distributions of all $\binom{5}{2}=10$ pairwise phase pairs, is reported in Table~\ref{tab:period_comparisons}; seven of ten contrasts are negligible, with the strongest effects clustered around the post-blackout transition mirroring the lexical-level drift pattern above.

\subsection{Topic Modeling: Extended Methodology and Results}
\label{app:topic_modeling}

\subsection*{Detailed Methodology}
\label{app:topic_methods}

To understand how the online conversation changed during the uprising, we follow a simple three-step process. First, a multilingual sentence transformer represents each comment so that comments with similar meanings are placed close together. Second, UMAP reduces these representations to a smaller space. Third, HDBSCAN groups nearby comments into topics. We also use a custom list of 619 Bangla stopwords, including 158 discourse markers, 151 platform-specific artifacts, and 6 special tokens. Removing this noise helps the model focus on meaningful themes. Table~\ref{tab:period_topics} reports the topic statistics for each event phase.

We choose the model settings to keep the topics detailed but still easy to interpret. For the global model, UMAP uses \texttt{n\_components=5}, \texttt{n\_neighbors=20}, and \texttt{min\_dist=0.1}. HDBSCAN uses \texttt{min\_cluster\_size=20}, \texttt{min\_samples=5}, and \texttt{cluster\_selection\_epsilon=0.1}. These lenient settings prevent the crisis discussion from being divided into too many small topics. CountVectorizer uses one- to three-word phrases, with \texttt{min\_df=10} and \texttt{max\_df=0.9}, to retain useful Bangla phrases while removing terms that are too rare or too common.

The five event phases contain different numbers of comments, so one clustering setting would not suit every phase. We therefore scale \texttt{min\_cluster\_size} from 25 in the smallest phase to 100 in the largest. If HDBSCAN marks a comment as an outlier (topic $=-1$), we assign it to the nearest topic centroid using cosine similarity. This gives every comment a topic. As a final check, we vary \texttt{min\_cluster\_size} by $\pm20\%$. Topic coherence changes by less than 0.05 ($\Delta C_v<0.05$), showing that the main findings remain stable under reasonable parameter changes.

\subsection*{Extended Results}
\label{app:topic_results}

\textbf{Attention rises and falls in sharp bursts.} The story begins with an online discussion that does not grow smoothly. Instead, public attention moves in sudden waves as events unfold. Seventy-five percent of the topics have a coefficient of variation above 1.0, meaning that their activity changes greatly relative to their average level. The largest burst appears on August~9, three days after the collapse, when Topic T1 reaches 1,063 posts in a single day. A topic remains active for 44.8 days on average. The dominant topic, T0, changes momentum 15 times and has a volatility value of 4,269.78. Together, these patterns show how quickly attention moved from one concern to another during the crisis.

\noindent\textbf{The themes change, but they remain connected.} Although attention is highly volatile, the discussion does not repeatedly start from zero. Across 100 topic transformations, the mean semantic similarity is 0.93. In other words, themes usually evolve from earlier themes rather than disappear and reappear in unrelated forms. For example, student-protest discussion (\bn{ছাত্রলীগের, আন্দোলন}) develops into narratives about state violence (\bn{পুলিশ, তোলা}) with a similarity of 0.902. Confrontation rhetoric later becomes celebratory discourse (\bn{জড়িত, হাসিমাখা}) with a similarity of 0.954, while its volume grows by 462\% after the collapse.

\noindent\textbf{Negative sentiment shapes most of the conversation.} The final part of the story concerns emotion. Only Topic T8 (\bn{সিদ্ধান্ত, সংস্কার}---decisions and reforms) has a positive polarity ($+0.246$). The most negative topics, T15 ($-0.486$) and T6 ($-0.294$), also carry a large share of the discussion. The topics are closely connected: we find 38 significant co-occurrence relationships with $r>0.5$. The strongest link is between decision-making and public-reaction topics (T3 $\leftrightarrow$ T1, $r=0.988$), reflecting the rapid cycle of action and reaction common in crisis communication~\cite{Tufekci2017}.

Taken together, the results tell a consistent story. Discussion stays centered on a small set of core grievances; these themes become more focused under stress, attract attention in sudden bursts, and continue to evolve across event phases. At the same time, sentiment remains mostly negative. Crisis discourse is therefore not simply a collection of disconnected conversations. It has a connected but rapidly changing structure, which calls for analysis methods designed specifically for crisis settings.

\subsection{Failure Mode Discovery: Topic-Level Sentiment Instability}

Overall performance scores hide an important part of the story: a model may appear stable on average while becoming unreliable for a particular topic. We therefore return to the 25 BERTopic topics and follow each one across the five event phases. For every topic, we measure how much its sentiment polarity changes using the standard deviation ($\sigma$). This topic-level view complements the lexical drift shown in Figure~\ref{fig:temporal_data_drift} and reveals that sentiment instability is concentrated in specific themes.

\begin{figure}[!t]
    \centering
    \includegraphics[width=\columnwidth]{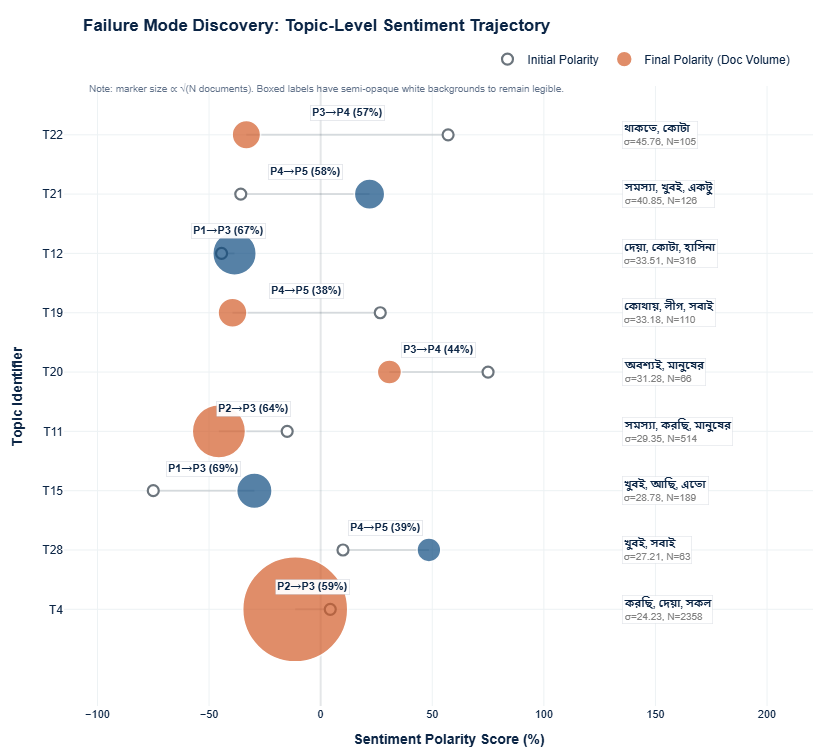}
    \caption{How topic-level sentiment changes across the five phases (P1--P5) of the July Revolution. A hollow circle shows a topic's initial polarity, and a filled circle shows its final polarity; larger filled circles represent more documents. Arrows and percentage labels show the direction and size of each change. T22 (\bn{কোটা}) is highly volatile, moving from $+57.1\%$ to $-33.3\%$ ($\sigma=45.76$), while flood-relief topics remain comparatively stable ($\sigma\approx12$--$15$). The largest reversals occur after the internet blackout (P2$\rightarrow$P3) and during the regime collapse (P3$\rightarrow$P4), showing that sentiment shift depends strongly on both topic and time.}
    \label{fig:topic_sentiment_trajectory}
    \vspace{-0.3cm}
\end{figure}

\noindent\textbf{Different topics follow different paths.} Six topics are highly volatile ($\sigma>30$), 14 have medium volatility ($15\leq\sigma\leq30$), and 5 remain stable ($\sigma<15$). Topic T22 (\bn{কোটা}---quota reform) provides the clearest example. Its polarity changes from $+57.1\%$ to $-33.3\%$ between P3 and P4, giving $\sigma=45.76$ and a range of 90.48 (Figure~\ref{fig:topic_sentiment_trajectory}). Humanitarian topics, in contrast, remain much steadier, with $\sigma\approx12$--15. The key lesson is that temporal shift does not affect every topic in the same way.

\noindent\textbf{Two turning points create the greatest risk.} Fifteen topics show a dramatic sentiment flip ($\Delta>40\%$), but most of the largest changes occur at two moments in the uprising. The first is the P2$\rightarrow$P3 transition, after the internet blackout, when information returns and people begin to interpret what happened. Four topics change by more than 60\% at this point. The second is the P3$\rightarrow$P4 transition, during the regime collapse, when celebration and uncertainty appear together. Five topics reverse by more than 45\%. For example, Topic T11 (\bn{সমস্যা}---problems) moves from $-15.0\%$ to $-45.6\%$ during P2$\rightarrow$P3 ($\Delta=64.2\%$). A classifier trained only on P2 would therefore underestimate P3 negativity for this topic by more than 30 percentage points.

\noindent\textbf{This pattern explains where models are likely to fail.} Politically charged themes, such as quota reform and league politics, are three times more volatile than humanitarian concerns. A single fixed model may therefore work well for stable topics but lose reliability when the meaning and sentiment of a political topic change. One practical response is topic-aware temporal recalibration: the model could lower or adjust its confidence when it detects a historically volatile topic. Future work could also prioritize these topics during temporal fine-tuning. Both strategies follow the same principle---adapt the model where the conversation changes most.

\subsection{Failure Mode Discovery through Topic Modeling}
\label{appendix:error_topic_modeling}

An overall accuracy score tells us how often a model is wrong, but it does not tell us whether the errors follow a common pattern. To find that pattern, we collect all 2,789 misclassified comments and apply BERTopic to the error set alone. Using a smaller configuration suited to this subset (\texttt{min\_cluster\_size=15} and one- to two-word phrases), the model finds 33 semantically coherent error topics with $C_v=0.9499$. We then follow the evidence from class-level errors to topic concentration, confidence, and complete sentiment reversals.

\noindent\textbf{The main problem begins with neutral comments.} The model classifies negative comments with 83.71\% accuracy and positive comments with 63.10\% accuracy, but neutral accuracy falls to 42.54\%. Thus, 57.46\% of neutral comments are misclassified. The largest single error type is neutral$\rightarrow$negative: it occurs 895 times and accounts for 32.1\% of all errors. This is the first sign that the model systematically reads ambiguous language as negative.

\noindent\textbf{The errors gather around a small set of themes.} They are not spread evenly across the 33 topics. Topic~0 alone contains 676 errors, or 24.2\% of the full error set. Among its 322 neutral errors, 261 (81.1\%) are predicted as negative. Large error clusters also appear around political discussion (\bn{বৈষম্যবিরোধী আন্দোলনের}) and education policy (\bn{আগের শিক্ষাক্রমে}). These examples suggest that BanglaBERT often reacts to words associated with conflict or criticism but misses the neutral pragmatic meaning of the full comment. LoRA-SFT on a mid-scale LLM recovers several such context-dependent cases, as illustrated in Table~\ref{tab:sft_vs_zeroshot}.

\begin{table}[t]
    \centering
    \caption{The ten topics containing the most classification errors. Topic~0 contains 676 errors (24.2\% of all errors), and 57.46\% of neutral comments are misclassified across the full error set. The mean confidence on incorrect predictions is 0.977, showing that the model is often highly confident even when it is wrong.}
    \label{tab:misclassification_topics}
    \resizebox{\columnwidth}{!}{%
    \begin{tabular}{lrcccc}
    \toprule
    \textbf{Topic} & \textbf{Errors} & \textbf{Conf.} & \textbf{Neu$\rightarrow$Neg} & \textbf{Neu$\rightarrow$Pos} & \textbf{Neg$\leftrightarrow$Pos} \\
    \midrule
    T0: \bn{আন্দোলনের স্পিরিটের}     & 676  & 0.968 & 261 & 61  & 104 \\
    T1: \bn{ঘটেছে বৈষম্যবিরোধী}       & 369  & 0.971 & 186 & 20  & 58 \\
    T2: \bn{বিদ্যুৎকেন্দ্র}              & 134  & 0.966 & 29  & 5   & 23 \\
    T3: \bn{আগের শিক্ষাক্রমে}          & 110  & 0.968 & 40  & 18  & 29 \\
    T4: \bn{শিক্ষার্থীদের স্বপ্নকে}    & 92   & 0.970 & 14  & 13  & 38 \\
    T5: \bn{আটক মির্জাপুরে}            & 89   & \textbf{1.000} & 0   & 85  & 4 \\
    T6: \bn{ছাত্রলীগের হামলায়}        & 75   & 0.980 & 40  & 7   & 10 \\
    T7: \bn{চিরকালই বিদ্রোহী}          & 75   & 0.982 & 19  & 9   & 17 \\
    T8: \bn{শিক্ষার্থীদের অস্বস্তি}    & 72   & 0.992 & 10  & 16  & 27 \\
    T9: \bn{দূর আগের}                   & 70   & 0.977 & 10  & 11  & 27 \\
    \midrule
    \textbf{All Topics (33)}            & 2{,}789 & 0.977 & 895 & 356 & 531 \\
    \bottomrule
    \end{tabular}%
    }
    \vspace{-0.3cm}
\end{table}

\noindent\textbf{The table confirms a consistent negative bias.} As Table~\ref{tab:misclassification_topics} shows, the model misclassifies 1{,}251 of the 2{,}177 neutral comments (57.46\%). It predicts negative for 895 of them and positive for 356, a ratio of about 2.5:1. The same direction of bias appears in T1, T3, and T6, where neutral$\rightarrow$negative errors clearly outnumber neutral$\rightarrow$positive errors. The problem therefore extends beyond one topic: ambiguous Bangla discourse is repeatedly pulled toward the negative class.

\noindent\textbf{High confidence makes these errors harder to detect.} The model's mean confidence across all incorrect predictions is 0.977. Topic~5 is the clearest warning: the confidence is 1.000 even though 85 neutral comments are predicted as positive. The model is therefore not simply uncertain about difficult examples; it is often confidently wrong. A standard probability threshold may fail to catch these cases, which motivates confidence calibration that also considers the topic.

\noindent\textbf{Some mistakes reverse the sentiment completely.} In 531 cases, or 19.0\% of all errors, the prediction flips directly between negative and positive. Topic~0 contains 104 such flips and Topic~1 contains 58; both focus on \bn{বৈষম্যবিরোধী আন্দোলনের}, where political framing can change the sentiment of otherwise similar language. Topic~4 (\bn{শিক্ষার্থীদের স্বপ্নকে}---students' dream) has an especially high flip rate: 38 of its 92 errors, or 41.3\%. The full error story is therefore not only about neutral comments. BanglaBERT also struggles to distinguish hopeful and critical framings when they discuss the same political or educational issue.

\subsection{Structural Diagnosis: Where Do Models Fail?}
\label{subsec:topic_modeling}

The previous section showed what the model's errors look like. We now ask why those errors gather in particular places. For this purpose, BERTopic~\cite{Grootendorst2022BERTopic} gives us two connected views. The first follows the topic structure of the full corpus over time; the second shows where the model's errors sit within that structure. The pipeline---multilingual sentence-transformer embeddings, UMAP, HDBSCAN, phase-specific parameters, and a 619-term Bangla stopword list---is described in Appendix~\ref{app:topic_methods}. Table~\ref{tab:period_topics} summarizes the results by phase. Two structural patterns explain much of the model behavior.

\noindent\textbf{First, one political theme dominates the conversation.} The global model finds 29 topics ($C_v=0.375$, silhouette $=0.402$), but T0 (\bn{লীগ, সরকার, কোটা}---league, government, quota) contains 56.3\% of all documents ($n=105{,}226$). It is also highly stable over time (CV $=0.014$) and has a negative polarity ($-0.224$). The effective number of topics based on Shannon entropy is only 1.7--2.0 across the five phases. Thus, the discussion may look diverse on the surface, but most attention remains concentrated around one central political grievance. We call this pattern \emph{discourse hegemony}.

This dominance creates a risk for sentiment models. Because T0 supplies such a large share of the training data, a model can learn to connect its political vocabulary with negative sentiment. It may then apply that negative tendency to neutral comments or to comments from other topics. This provides a structural explanation for the neutral$\rightarrow$negative bias found in Appendix~\ref{appendix:error_topic_modeling}.

\noindent\textbf{Second, the blackout makes the discussion more focused, not more random.} Topic coherence is not determined by the number of comments. P2, the Crisis \& Blackout phase, contains 33,684 comments (16.9\% of the corpus) and produces 29 topics, yet it has the highest coherence by a wide margin ($C_v=0.781$). Coherence in every other phase ranges only from 0.384 to 0.422. The largest phase, P4 (Post-Revolution), contains 69,536 comments (34.9\%) but reaches only $C_v=0.420$. The smallest phase, P1 (Pre-Escalation), contains 6,727 comments (3.4\%) and reaches $C_v=0.405$.

We call this result the \emph{crisis coherence paradox}. During the blackout, limited information and immediate danger push people toward a small set of urgent concerns, making the discussion more semantically focused. After the crisis, attention spreads across more issues and coherence falls. Extreme conditions therefore crystallize the conversation instead of turning it into noise.

\noindent\textbf{Together, these patterns explain where models fail.} Errors cluster around the dominant topic and closely related political themes rather than appearing uniformly across the corpus. Politically charged topics, including quota reform and party politics, show three times the sentiment volatility of humanitarian topics, and 33\% of all errors fall within T0's thematic orbit. Temporal robustness should therefore be measured separately for each topic. In the same way, confidence recalibration should use topic-level volatility instead of relying only on one aggregate confidence score.

\subsection{Prompting Techniques}
\label{appendix:prompting_techniques}

After identifying where sentiment models struggle, we test whether LLMs can solve the task from instructions and examples alone. We compare two prompting settings: zero-shot and few-shot. The model parameters remain frozen in both settings, so the only difference is the information supplied in the prompt. This design separates in-context reasoning from any benefit that could come from task-specific fine-tuning.

\noindent\textbf{Zero-shot prompting.}
We begin with the harder setting: the model receives no labeled examples. The prompt contains only a short expert-role instruction (for example, ``You are an expert in Bengali political sentiment analysis''), the parent \texttt{post}, the target \texttt{comment}, the allowed labels $\labels$, and an instruction to return one label. The resulting score shows how well the model can understand crisis-era Bangla using its pretrained knowledge alone. Success therefore depends on its Bangla vocabulary, cultural and political understanding, and ability to follow instructions.

\noindent\textbf{Few-shot prompting.}
Next, we give the same model seven labeled examples from the training split of \dataset{}. Each example contains a \texttt{post}, a \texttt{comment}, and the correct \texttt{label}. Together, the examples cover all three sentiment classes, both supportive and critical views of crisis events, and difficult language such as sarcasm, rhetorical questions, emoji descriptions, Bangla--English code-mixing, and indirect political references. The model then labels a held-out post--comment pair under the same one-label rule. Comparing this result with zero-shot performance shows how much seven relevant examples can help without changing the model itself.

\noindent\textbf{Decoding and output format.}
Finally, we keep the decision process identical across both settings. We use greedy decoding with temperature $=0$ and require exactly one label from $\labels$. If a response does not match an allowed label, we count it as an error and do not ask the model again. The reported results therefore measure the reliability of the complete prompting process, including instruction following, rather than the quality of predictions after manual correction or extra parsing. The complete prompt templates appear in~\S\ref{appendix:sample_prompt_result}.

\subsection{Expert Prompting with LLMs}
\label{appendix:expert_prompting}

Table~\ref{tab:expert_bn} reports the zero-shot and few-shot performance of open-source and proprietary LLMs using expert prompting with Bangla-native instructions on the UnRestSent200K benchmark.

\section{Ethics Statement}
\label{sec:ethics}

All data used in this work were collected from publicly accessible Facebook and YouTube content in accordance with platform terms of service. Personally identifiable information, including usernames and profile links, was removed during preprocessing.

\subsection{Sample Prompts}
\label{appendix:sample_prompt_result}
\vspace{-2mm}

\onecolumn
\raggedbottom

\vspace{2mm}

\begin{promptbox}[\bn{ফিউ-শট প্রম্পট: সেন্টিমেন্ট অ্যানোটেশন} (Few-Shot: Sentiment Annotation --- Bengali)]

\textbf{System:} \bn{তুমি বাংলা ভাষার রাজনৈতিক sentiment analysis বিশেষজ্ঞ। তোমাকে একটি সংবাদ পোস্ট/শিরোনাম এবং সেই পোস্টের উপর করা একটি মন্তব্য দেওয়া হবে। তোমার কাজ হলো নির্ধারণ করা যে মন্তব্যটি উল্লিখিত পোস্ট, ঘটনা, বা প্রসঙ্গের প্রতি কী ধরনের sentiment প্রকাশ করছে।}

\vspace{1mm}
\textbf{User:} \bn{নিচে কিছু উদাহরণ দেওয়া আছে। উদাহরণগুলো অনুসরণ করে শেষে দেওয়া প্রশ্নটির উত্তর দাও।}

\vspace{1mm}
\bn{নির্দেশনা:}
\begin{itemize}[leftmargin=*, nosep]
\item \bn{post শুধুমাত্র contextual information হিসেবে ব্যবহার করবে।}
\item \bn{sentiment নির্ধারণের সময় sarcasm, rhetorical tone, slang, emoji description (যেমন: `কান্না জড়িত হাসিমুখ প্রতিক্রিয়া'), এবং mixed Bangla-English expression বিবেচনা করবে।}
\item \bn{রাজনৈতিক মতাদর্শ বা ব্যক্তিগত অবস্থান নয়, শুধুমাত্র comment-এর sentiment বিচার করবে।}
\item \bn{যদি comment সমর্থন, প্রশংসা, আশাবাদ, বা ইতিবাচক প্রতিক্রিয়া প্রকাশ করে} $\rightarrow$ \texttt{positive}
\item \bn{যদি comment সমালোচনা, হতাশা, রাগ, ব্যঙ্গ, আক্রমণ, বা নেতিবাচক প্রতিক্রিয়া প্রকাশ করে} $\rightarrow$ \texttt{negative}
\item \bn{যদি comment তথ্যভিত্তিক, প্রশ্নধর্মী, দ্ব্যর্থক, বা স্পষ্ট ইতিবাচক/নেতিবাচক sentiment না প্রকাশ করে} $\rightarrow$ \texttt{neutral}
\end{itemize}

\vspace{1mm}
\bn{উদাহরণসমূহ:}

\textbf{\bn{উদাহরণ ১}}\\
\bn{পোস্ট: কোটাবিরোধী আন্দোলন : ঢাকা বিশ্ববিদ্যালয়ে মিছিল শুরু, মধুর ক্যানটিনে জড়ো হয়েছে ছাত্রলীগ}\\
\bn{মন্তব্য: সাবধান, আবার রক্ত ঝরাবে।}\\
\bn{উত্তর:} \texttt{negative}

\textbf{\bn{উদাহরণ ২}}\\
\bn{পোস্ট: অপরাধে সম্পৃক্ততায় যুবদল--ছাত্রদলের ১৫ জনকে বহিষ্কার}\\
\bn{মন্তব্য: নাটক কম করো পিও, জনগণ সব বুঝে।}\\
\bn{উত্তর:} \texttt{negative}

\textbf{\bn{উদাহরণ ৩}}\\
\bn{পোস্ট: আজ বৃহস্পতিবার দেশের প্রায় সব মোবাইল অপারেটরের ইন্টারনেট ধীরগতিতে চলছে।}\\
\bn{মন্তব্য: শ্লা টাউট! কান্না জড়িত হাসিমুখ প্রতিক্রিয়া}\\
\bn{উত্তর:} \texttt{negative}

\textbf{\bn{উদাহরণ ৪}}\\
\bn{পোস্ট: কোটাবিরোধী আন্দোলন : ঢাকা বিশ্ববিদ্যালয়ে মিছিল শুরু, মধুর ক্যানটিনে জড়ো হয়েছে ছাত্রলীগ}\\
\bn{মন্তব্য: ছাত্রলীগও কোটা আন্দোলনে অংশ নিয়েছে? ভাবনা মুখ প্রতিক্রিয়া}\\
\bn{উত্তর:} \texttt{neutral}

\textbf{\bn{উদাহরণ ৫}}\\
\bn{পোস্ট: অপরাধে সম্পৃক্ততায় যুবদল--ছাত্রদলের ১৫ জনকে বহিষ্কার}\\
\bn{মন্তব্য: যাদের পদ নেই তাদের কি শাস্তি হবে?}\\
\bn{উত্তর:} \texttt{neutral}

\textbf{\bn{উদাহরণ ৬}}\\
\bn{পোস্ট: কোটাবিরোধী আন্দোলন : ঢাকা বিশ্ববিদ্যালয়ে মিছিল শুরু, মধুর ক্যানটিনে জড়ো হয়েছে ছাত্রলীগ}\\
\bn{মন্তব্য: কোটা সিস্টেম নিপাত যাক। মেধাবীরা চাকরি পাক।}\\
\bn{উত্তর:} \texttt{positive}

\textbf{\bn{উদাহরণ ৭}}\\
\bn{পোস্ট: অপরাধে সম্পৃক্ততায় যুবদল--ছাত্রদলের ১৫ জনকে বহিষ্কার}\\
\bn{মন্তব্য: সঠিক সময়ের সঠিক সিদ্ধান্ত ধন্যবাদ}\\
\bn{উত্তর:} \texttt{positive}

\vspace{1mm}
\bn{এখন নিচের উদাহরণটির sentiment নির্ধারণ করো:}

\bn{পোস্ট:} \texttt{\{post\}}

\bn{মন্তব্য:} \texttt{\{comment\}}

\vspace{1mm}
\bn{শুধুমাত্র নিচের একটি label প্রদান করো:} \texttt{positive}, \texttt{neutral}, \texttt{negative}

\end{promptbox}

\vspace{2mm}

\begin{promptbox}[Few-Shot: Sentiment Annotation --- English]

\textbf{System:} You are an expert in political sentiment analysis for the Bangla language. You will be given a news post or headline and a comment made on that post. Your task is to determine what kind of sentiment the comment expresses toward the mentioned post, event, or context.

\vspace{1mm}
\textbf{User:} Some examples are given below. Follow the examples and answer the final question.

\vspace{1mm}
\textbf{Instructions:}
\begin{itemize}[leftmargin=*, nosep]
\item Use the post only as contextual information.
\item While determining sentiment, consider sarcasm, rhetorical tone, slang, emoji descriptions such as ``crying laughing face reaction'', and mixed Bangla-English expressions.
\item Judge only the sentiment of the comment, not political ideology or personal position.
\item If the comment expresses support, praise, optimism, or a positive reaction, label it \texttt{positive}.
\item If the comment expresses criticism, frustration, anger, sarcasm, attack, or a negative reaction, label it \texttt{negative}.
\item If the comment is factual, question-like, ambiguous, or does not express a clear positive or negative sentiment, label it \texttt{neutral}.
\end{itemize}

\vspace{1mm}
\textbf{Examples:}

\textbf{Example 1}\\
Post: Anti-quota movement: A procession has started at Dhaka University, and Chhatra League has gathered at Madhur Canteen.\\
Comment: Be careful, they will shed blood again.\\
Answer: \texttt{negative}

\textbf{Example 2}\\
Post: Fifteen members of Jubo Dal and Chhatra Dal expelled for involvement in crimes.\\
Comment: Stop the drama, people understand everything.\\
Answer: \texttt{negative}

\textbf{Example 3}\\
Post: Today, Thursday, internet service is slow across almost all mobile operators in the country.\\
Comment: Damn fraud! Crying laughing face reaction.\\
Answer: \texttt{negative}

\textbf{Example 4}\\
Post: Anti-quota movement: A procession has started at Dhaka University, and Chhatra League has gathered at Madhur Canteen.\\
Comment: Has Chhatra League also joined the quota movement? Thinking face reaction.\\
Answer: \texttt{neutral}

\textbf{Example 5}\\
Post: Fifteen members of Jubo Dal and Chhatra Dal expelled for involvement in crimes.\\
Comment: Will those who do not hold any position also be punished?\\
Answer: \texttt{neutral}

\textbf{Example 6}\\
Post: Anti-quota movement: A procession has started at Dhaka University, and Chhatra League has gathered at Madhur Canteen.\\
Comment: Down with the quota system. Let meritorious students get jobs.\\
Answer: \texttt{positive}

\textbf{Example 7}\\
Post: Fifteen members of Jubo Dal and Chhatra Dal expelled for involvement in crimes.\\
Comment: The right decision at the right time. Thank you.\\
Answer: \texttt{positive}

\vspace{1mm}
\textbf{Now determine the sentiment of the following example:}

\textbf{Post:} \texttt{\{post\}}

\textbf{Comment:} \texttt{\{comment\}}

\vspace{1mm}
Provide only one of the following labels: \texttt{positive}, \texttt{neutral}, or \texttt{negative}.

\end{promptbox}

\vspace{2mm}
\begin{promptbox}

\textbf{Zero-Shot Prompt: Political Expert (Bengali)}

\bn{তুমি একজন বাংলা রাজনৈতিক sentiment analysis বিশেষজ্ঞ। নিচের বাংলা পোস্ট এবং মন্তব্যটির sentiment নির্ধারণ করো।}

\vspace{1mm}

\bn{পোস্ট:} \verb|{post}|

\bn{মন্তব্য:} \verb|{comment}|

\vspace{1mm}

\bn{শুধুমাত্র একটি লেবেল দাও:} \texttt{positive}, \texttt{neutral}, \texttt{negative}।

\end{promptbox}

\vspace{2mm}

\begin{promptbox}[Zero-Shot: Political Expert --- English]

You are an expert in Bengali political sentiment analysis. Determine the sentiment of the following Bengali post and comment.

\vspace{1mm}
\textbf{Post:} \texttt{\{post\}}

\textbf{Comment:} \texttt{\{comment\}}

\vspace{1mm}
Respond with only one label: \texttt{positive}, \texttt{neutral}, or \texttt{negative}.

\end{promptbox}

\end{document}